\documentclass[a4paper, 10pt, conference]{ieeeconf}      %
\usepackage{FG2026}

\usepackage[dvipsnames]{xcolor}
\usepackage{pifont}
\usepackage{multirow}
\usepackage{colortbl}
\usepackage{comment}
\usepackage{relsize}
\usepackage{dsfont}
\usepackage{xspace}
\usepackage{amsmath}
\usepackage{graphicx}
\usepackage{wrapfig}
\usepackage{lipsum}
\usepackage{bm}
\usepackage{booktabs}
\usepackage{amssymb}
\usepackage{hyperref}
\usepackage[capitalize]{cleveref}
\crefname{section}{Sec.}{Secs.}
\Crefname{section}{Section}{Sections}
\Crefname{table}{Table}{Tables}
\crefname{table}{Tab.}{Tabs.}

\let\titleold\title
\renewcommand{\title}[1]{\titleold{#1}\newcommand{\thetitle}{#1}}
\def\maketitlesupplementary
{
\newpage
   \twocolumn[
    \centering
    \Large
    \textbf{\thetitle}\\
    \vspace{0.5em}Supplementary Material \\
    \vspace{1.0em}
   ] %
}

\def\eg{\emph{e.g}\xspace} 
\def\ie{\emph{i.e}\xspace}

\definecolor{ourtablecolor}{RGB}{235, 235, 235}
\definecolor{goodgreen}{rgb}{0.0, 0.56, 0.0}
\newcommand{\gooddelta}[1]{\scalebox{1.05}{$_{\textcolor{goodgreen}{\scriptscriptstyle\uparrow\bm{#1}}}$}}

\newcommand{\method}{AnyHOI\xspace}
\newcommand{\task}{U-HOI\xspace}
\newcommand{\taskacro}{Unconstrained HOI\xspace}
\newcommand{\publicsetting}{annotated-box\xspace}
\newcommand{\privatesetting}{computed-box\xspace}
\newcommand{\lmm}{MLLM\xspace}
\newcommand{\lmms}{MLLMs\xspace}

\newcommand{\hicodet}{HICO-DET~\cite{Chao2018}\xspace}
\newcommand{\vghoi}{VG-HOI~\cite{Wu2024}\xspace}
\newcommand{\supmat}{Supp. Mat.\xspace}
\newcommand{\cmark}{\textcolor{ForestGreen}{\ding{51}}}
\newcommand{\xmark}{\textcolor{red}{\ding{55}}}

\FGfinalcopy %

\IEEEoverridecommandlockouts                              %
\def\FGPaperID{XXX} %

\title{\LARGE \bf
Towards Unconstrained Human-Object Interaction
}

\author{\parbox{16cm}{\centering
    {\large Francesco Tonini$^{1, 2}$, Alessandro Conti$^1$, Lorenzo Vaquero$^2$, Cigdem Beyan$^3$, and Elisa Ricci$^{1,2}$}\\
    {\normalsize
    $^1$Department of Information Engineering and Computer Science, University of Trento, Trento, Italy\\
    $^2$Fondazione Bruno Kessler, Trento, Italy, \\
    $^3$Department of Computer Science, University of Verona, Verona, Italy}}
}

\usepackage{fancyhdr}
\begin{document}

\ifFGfinal
\thispagestyle{empty}
\pagestyle{empty}
\else
\author{Anonymous FG2026 submission\\ Paper ID \FGPaperID \\}
\pagestyle{plain}
\fi
\maketitle
\thispagestyle{fancy}
\renewcommand{\headrulewidth}{0pt}
\fancyhf{}
\fancyhead[C]{2026 International Conference on Automatic Face and Gesture Recognition (FG)}

\begin{abstract}
Human-Object Interaction (HOI) detection is a longstanding computer vision problem concerned with predicting the interaction between humans and objects. Current HOI models rely on a vocabulary of interactions at training and inference time, limiting their applicability to static environments. With the advent of Multimodal Large Language Models (\lmms), it has become feasible to explore more flexible paradigms for interaction recognition. In this work, we revisit HOI detection through the lens of \lmms and apply them to in-the-wild HOI detection. We define the \taskacro (\task) task, a novel HOI domain that removes the requirement for a predefined list of interactions at both training and inference. We evaluate a range of \lmms on this setting and introduce a pipeline that includes test-time inference and language-to-graph conversion to extract structured interactions from free-form text. Our findings highlight the limitations of current HOI detectors and the value of \lmms for \task. Code will be available at: \url{https://github.com/francescotonini/anyhoi}
\end{abstract}
    
\section{Introduction}
\label{sec:intro}
Human-Object Interaction (HOI) detection is a fundamental problem in computer vision that requires simultaneously localizing humans and objects in images while identifying their interactions. %
Given an input image, HOI detectors identify human-object pairs and classify their interactions as $\langle \text{\textit{human}}, \text{\textit{verb}}, \text{\textit{object}} \rangle$ triplets.

Traditional approaches rely on a fixed set of manually defined and labeled interaction categories, which are learned during training and remain unchanged at test time (\textit{closed-set} setting)~\cite{zhang2023exploring,zhang2022exploring,Zhang2022,kim2021hotr}.
These models require large, labeled datasets like \hicodet, in which every interaction is explicitly specified.
While this approach works well in controlled, data-rich environments, it inherently limits the model's ability to recognize interactions beyond those seen during training, restricting its applicability in dynamic real-world scenarios.
Open-vocabulary~\cite{Wu2024,Yang24,Lei2023,Mao2023,Ning2023,kim2025locality} HOI detectors overcome these limitations by integrating and finetuning large Vision-Language Models (VLMs) such as CLIP~\cite{Radford2021} and BLIP~\cite{li2023blip}.
Typically, these methods are trained on a base set of interactions and test to unseen interactions during inference.
However, finetuning models on a subset of interactions penalizes predictions towards unseen and rare interactions~\cite{guo2024unseen}, and
\emph{it still depends on a fixed, predetermined interaction vocabulary at inference time}.
Therefore, these models cannot detect interactions beyond the provided vocabulary, making them unsuited to in-the-wild scenarios where enumerating all possible human-object interactions apriori is impractical, as in HOI models applied to dynamic environments such as assistive robotics~\cite{mascarohoi4abot,alameda2025socially}, autonomous driving~\cite{chen2024asynchronous}, in-the-wild human intention understanding~\cite{zunino2020predicting}, or activity parsing~\cite{luo2021moma}.

\begin{figure}[!t]
\centering
\includegraphics[width=0.9\linewidth]{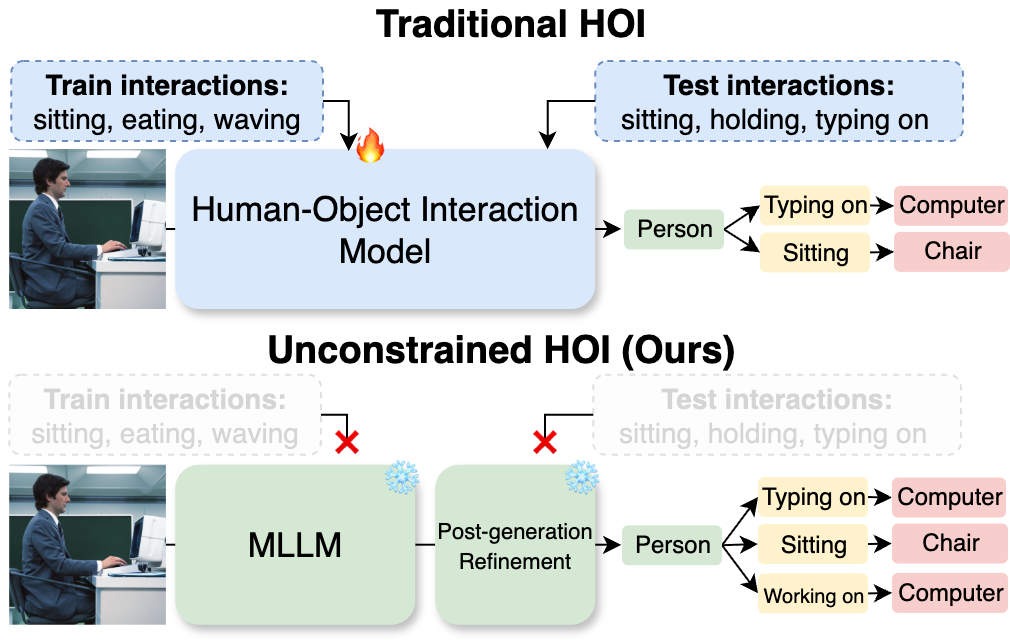}
\caption{
Comparison between traditional HOI settings and our \task.
Unlike prior art, \task does not require training and eliminates the need for a predefined list of interactions.
We tackle \task using off-the-shelf Multimodal Large Language Models (\lmms), processing their free-form outputs and applying a post-generation refinement that extracts HOI triplets, all without being constrained by a fixed vocabulary.}
\label{fig:teaser}
\end{figure}

For example, imagine a humanoid robot assisting an elderly person at home. One day, it encounters a newly delivered package: the person asks the robot to ``open the box'' and later to ``place the charger inside the drawer''. The next day, the same robot may need to interpret a different request, such as ``hand me the remote'' or ``help me fold the blanket.''.
None of these interactions can be exhaustively enumerated in advance, since the set of objects in a household is constantly changing and the actions people use are highly variable. A similar challenge arises in autonomous driving, where a car might need to interpret rare but safety-critical interactions like a pedestrian ``pushing a stroller,'' or ``waving to stop the vehicle.'' In healthcare or rehabilitation, patients often improvise with unfamiliar objects or perform nonstandard movements that lie outside any predefined interaction list. 
Real-world HOI detection must operate in open, unpredictable environments, where constraining the system to a fixed set of verb–noun pairs is unrealistic.

In this paper, we challenge 
the reliance on a predefined interaction vocabulary and propose a new task: \textbf{\taskacro (\task)}.
The goal of \task is to  
recognize arbitrary human-object interactions without relying on a fixed set of interaction labels, both during training and inference.
By removing such constraints, \task forces HOI models to identify the underlying HOI patterns in an open-ended manner, enabling them to detect interactions beyond predefined categories.
To evaluate HOI detection in this previously unexplored setting, we introduce \emph{Semantic Recall (SR)}, which, alongside mAP, provides a comprehensive analysis of the interactions retrieved by the model.

To address the \task task, we compare several off-the-shelf Multimodal Large Language Models (\lmms) leveraging their perception capabilities, and introduce \method, a training-free reference pipeline for studying \task that extracts human-object interaction triplets from \lmms.
Unlike other \lmm-based methods that require a fixed interaction vocabulary~\cite{Xie25}, our approach exploits the strong representational power and generative nature of \lmms to produce 
free-form
textual descriptions of scenes (Fig.~\ref{fig:teaser}).
These generations may contain unrelated information (\eg, objects' attributes)
limiting their direct parsing to HOI triplets (see \cref{sec:experiments-benchmarking}).
To overcome this challenge, 
we refine the \lmm outputs with a post-generation pipeline that constructs scene graphs, enabling the removal of irrelevant relationships and the generation of HOI triplets.
Furthermore, 
encouraged by the results in mathematical reasoning~\cite{hendrycks2021measuring} and visual understanding~\cite{liu2024mmbench}, we employ a test-time compute strategy for \task, so as to improve the reasoning capabilities of smaller \lmms and increase their recall. 
We provide an initial diagnostic tool for studying the \task task that can be readily applied to a wide range of \lmms.
This allows us to bridge the gap between the free-form outputs of \lmms and the structured triplet format required for HOI detection and downstream applications, such as assistive technologies~\cite{mascarohoi4abot,chen2024asynchronous} and human behavior prediction~\cite{zunino2020predicting, luo2021moma}.

We emphasize that \task is not meant to replace closed-set or open-vocabulary HOI detection, but to introduce a setting where no predefined interaction vocabulary is assumed at inference. This better reflects real-world scenarios where enumerating all possible human–object interactions is impractical. The main \textbf{contributions} of our work are as follows:
\begin{itemize}
    \item We formalize the \taskacro (\task) task, a novel setting for in-the-wild HOI detection that removes the requirement for a fixed interaction vocabulary at both training and inference time, and introduce an evaluation protocol with newly defined metrics.
    \item We conduct a diagnostic analysis of existing open-vocabulary HOI detectors and \lmms on \task, revealing their capabilities and limitations under this challenging, deployment-oriented setting.
    \item We introduce \method, a training-free reference pipeline for studying \task that combines \lmms with post-generation refinement for triplet extraction and a test-time compute strategy for improved interaction detection.
\end{itemize}

\section{Related work}
\label{sec:related}
\noindent \textbf{Human-Object Interaction Detection.}
HOI detectors are usually classified as one-stage or two-stage. One-stage detectors handle object detection, pairing, and interaction classification simultaneously. Early variants rely on interaction points~\cite{Liao2020,Wang2020} and merged bounding-boxes~\cite{Kim2020a} to generate the candidate regions. Recent methods use DETR~\cite{Carion2020} to learn interaction queries, which are decoded into triplets. Despite their strong performance in fully-supervised settings, one-stage approaches tend to be computationally intensive and slow to converge.
Instead, two-stage methods
first propose human and object regions using off-the-shelf object detectors~\cite{Carion2020,Ren2017} and then extract interaction cues based on location~\cite{Ulutan2020,Gao2020,Zhang2021}, vision~\cite{Zhang2022}, pose information~\cite{Li2022a,Li2020,Wu2024b}, or occurrence priors~\cite{Kim2020b}.
Subsequent relation prediction is often implemented via stream fusion techniques~\cite{Chao2018,Gao2020} or message-passing~\cite{Ulutan2020,Zhang2021,Liu2020b}.
However, conventional supervised HOI detectors struggle with long-tailed distributions, leading to poor generalization on rare interaction classes~\cite{Lei2023,Mao2023}.
To address this, recent advances leverage VLMs to enhance zero-shot generalization and expand beyond the training set of interaction classes.

\noindent \textbf{VLMs for HOI.}
Vision-language models, such as CLIP~\cite{Radford2021} and BLIP~\cite{li2023blip}, leverage large-scale image-text pretraining to acquire a rich and generalizable understanding of visual concepts.
This capability has facilitated the development of open-vocabulary HOI detection methods, which identify interaction categories based on a user-provided verb list at inference time.
Some studies distill knowledge from VLMs~\cite{Liao2022,Wu2023,Wang2022b,Zhao2023}, while others replace the traditional feature extractors with VLM-based encoders~\cite{Park2023,Ning2023}.
Furthermore, prompting techniques are employed to extract additional HOI cues from scenes~\cite{Mao2023,Cao2023,Luo2024} and visual memory banks~\cite{Lei2023,tonini2025dynamic}, and interaction-aware calibration priors~\cite{Yang2024} are introduced to improve modality alignment.
Recent efforts benchmarks HOI detection~\cite{Wu2024} with out-of-domain interactions during testing, showing the limitations of HOI detections under such conditions.
However, a 
fundamental limitation 
of these existing approaches is that
they still require a predefined set of triplets at inference time.
We argue that such a constraint is unnecessary for real-world deployment scenarios.
\lmms can offer the potential to enable HOI detection in an open-ended semantic space, moving beyond fixed verb lexicons and allowing a more flexible and comprehensive understanding of HOI.

\noindent \textbf{\lmms for HOI.}
The rapid advancement of \lmms \cite{Liu23,wang2024qwen2,laurenccon2025matters} has opened promising avenues for rethinking HOI detection.
Vision-language instruction tuning~\cite{huang2023visual,li2024llava} has enabled the creation of general-purpose \lmms that can be adapted to a wide range of tasks through natural language interfaces.
Recently, some models have been explored for HOI detection~\cite{Xie25}, although they still adhere to the constraints of closed interaction vocabularies.

Achieving a truly unbounded HOI detection paradigm presents significant challenges.
While \lmms demonstrate enhanced semantic understanding and robustness to out-of-domain variations, they often lack the fine-grained verb comprehension required to accurately capture the dynamics of HOIs~\cite{Xie25,Momeni2023}.
Furthermore, their predictions can be unreliable (see \cref{tab:main_hicodet_classification}), highlighting the need for further research.
To bridge the gap between the representational power of \lmms and the specific demands of HOI detection tasks, we introduce the \task setting.

\section{\taskacro}
\label{sec:verb_free_hoi}
\begin{table}[!t]
\caption{Unlike conventional methods with predefined object-verb vocabularies, our \taskacro task infers interactions without prior combinations.
}
\centering
\resizebox{1.0\columnwidth}{!}{%
\begin{tabular}{@{}l|c@{\hskip 4pt}c@{\hskip 4pt}c@{\hskip 12pt}c@{}}
\toprule
& \multicolumn{3}{c}{\textbf{Interaction Vocabulary}}  &  \textbf{Object-Verb Priors} \\[-2pt]
\midrule
\textbf{Closed-set}, \eg,~\cite{Zhang2021,Kim2020a,Zhang2022} & \multicolumn{3}{c}{$\mathcal{V}_{train} = \mathcal{V}_{test}$} & Required \\
\textbf{Open-vocabulary}, \eg,~\cite{Lei2023,Mao2023,Ning2023,Yang24,Wu2024} & \multicolumn{3}{c}{$\mathcal{V}_{train} \neq \mathcal{V}_{test}$} & Required \\
\textbf{\taskacro (Ours)} & \multicolumn{3}{c}{Not provided} & Not required \\
\bottomrule
\end{tabular}%
}
\label{tab:settings_comparison}
\end{table}

\taskacro (\task) is a novel task formulation of HOI detection that eliminates any constraints and priors imposed by predefined 
vocabularies.
In \cref{sec:verb_free_hoi-formulation}, we formally define the problem setting and, in \cref{sec:verb_free_hoi-formulation}, we outline the evaluation metrics and protocol.

\subsection{Problem formulation}
\label{sec:verb_free_hoi-formulation}
\noindent \textbf{Preliminaries.}
Given an input image $\mathcal{I}$, HOI detection aims to predict a set of $\langle \text{\textit{human}}, \text{\textit{verb}}, \text{\textit{object}} \rangle$ triplets describing the interactions between the people and objects in the scene.
Traditionally, the object classes $\mathcal{O}$ and verb categories $\mathcal{V} \subset \Omega$, sampled from an unbounded vocabulary $\Omega$, are predefined for both training and testing.
This information is often encoded using a sparse co-occurrence matrix $\mathcal{M} \in \{0,1\}^{|\mathcal{O}| \times |\mathcal{V}|}$, which defines the set of verb-object interactions present in a given dataset.
For example, $\mathcal{M}$ might allow the triplet $\langle \text{\textit{person}}, \text{\textit{lying on}}, \text{\textit{bed}} \rangle$, but not $\langle \text{\textit{person}}, \text{\textit{sitting on}}, \text{\textit{bed}} \rangle$, as the latter interaction does not appear in the dataset~\cite{Chao2018}. 

\begin{figure*}[!ht]
    \centering
    \includegraphics[width=0.9\linewidth]{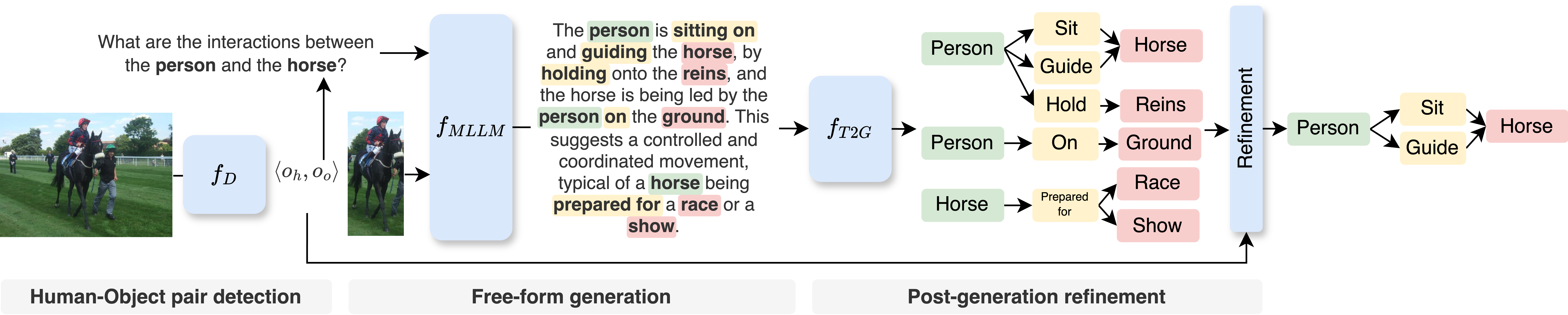}
    \caption{\textbf{\method} first detects human-object pairs using an object detector. Then, it crops the region encompassing both the human and its paired object and feeds it into an \lmm, along with a prompt to generate free-form scene descriptions. A post-generation refinement step then analyzes these descriptions to produce a relationship graph. Finally, the graph is refined by filtering out the triplets that are irrelevant.}
    \vspace{-1.5em}
    \label{fig:method}
\end{figure*}

Current HOI detectors, including open-vocabulary~\cite{Lei2023,Mao2023,Ning2023,Wu2024,Yang2024}, 
rely on both $\mathcal{V}$ and $\mathcal{M}$ during inference, as summarized in \cref{tab:settings_comparison}, to the point where they are unable to function if $\mathcal{M}$ or 
$\mathcal{V}$ are undefined (see \cref{sec:experiments}).
To overcome these limitations in current HOI literature, we propose a novel task that does not need fixed verb lexicons, thereby enabling HOI detection over an open-ended semantic space.

\noindent \textbf{Task definition.}
We propose \taskacro~(\task), a new HOI detection setting that requires predicting interactions without any prior knowledge of $\mathcal{V}$ or $\mathcal{M}$.
Unlike traditional approaches that operate within a predefined verb set, \task explores the full verb vocabulary space $\Omega$, making no assumptions about the underlying interaction distribution.
The primary challenge of \task stems from the vast and unbounded nature of the verb space ($|\Omega| \gg |\mathcal{V}|$).
To put this into perspective, natural language resources such as WordNet~\cite{miller1995wordnet} provide structured verb lexicons that, despite being smaller than $\Omega$, are several orders of magnitude larger than those found in standard HOI datasets, like \hicodet.
This extreme vocabulary expansion not only complicates interaction modeling but also raises fundamental questions about how to fairly and effectively evaluate HOI predictions in an open-ended setting.
In the following section, we introduce our evaluation metrics designed specifically for \task.

\subsection{Evaluation metrics}
\label{sec:verb_free_hoi-metrics}
In prior art~\cite{Xie25,Zhang2022,Ning2023,Wu2024}, where the predicted verb categories are restricted to a fixed vocabulary $\mathcal{V}$,
the performance of HOI models is evaluated with the mean Average Precision (mAP) across all categories in $\mathcal{V}$ (more details in \cref{sec:map}).
While mAP proved to be effective in closed-set and open-vocabulary HOI, where the verb categories are given and the model cannot predict verbs outside of those categories, it is not suitable for \task.
In this setting, models generate free-form interaction descriptions that are not constrained to a predefined verb vocabulary. As a result, exact lexical matching between predicted and ground-truth verbs becomes unreliable: semantically equivalent interactions may be expressed using different surface forms (e.g., ``sitting on'' vs. ``seated on''), while partially correct predictions may still capture relevant interaction intent.
Thus, we adapt mAP to reflect the new challenges of \task as well as introduce a new metric, \textit{Semantic Recall (SR)}, to better assess the capability of models in \task.

\noindent The \textbf{mean Average Precision (mAP)} remains largely unchanged.
Since predictions in \task are free-form, we map them back into the evaluation vocabulary via semantic similarity.
This allows us to calculate mAP as in the HOI literature, ensuring a fair evaluation with prior art.
Formally, let $\mathcal{K}' = \{(b_{h_i}, b_{o_i}, v'_i)\}_{i=1}^{N'}$ denote the set of predicted HOI triplets for an image, where $b_{h_i} \in \mathbb{R}^4$ and $b_{o_i} \in \mathbb{R}^4$ denote the human and object bounding boxes, respectively, and $v'_i \in \Omega$ is the interaction verb.
To match these unconstrained predictions with the verb evaluation set~$\mathcal{V}$, we follow the approach of~\cite{li2023factual, diko2024semantically} and employ a BERT-based~\cite{bert} similarity function that measures the semantic similarity between a predicted verb $v'_i$ and verbs $v \in \mathcal{V}$, and match predictions above a certain threshold $\tau$.

\noindent \textbf{Semantic Recall (SR)} 
measures whether a model is able to recover the semantic intent of a ground-truth interaction for a given human-object pair, regardless of the exact phrasing used. For each ground-truth interaction, SR evaluates the most semantically aligned prediction among among valid predictions.
Formally, given a set of ground-truth interactions  $\mathcal{G} =\{(\hat{b}_{h_g}, \hat{b}_{o_g}, \hat{v}_g)\}_{g=1}^{M}$, we first identify the set of predictions that spatially correspond to each ground-truth instance:
\begin{equation}
    \mathcal{K}'_g = \Big\{ (b_{h_i}, b_{o_i}, v'_i) \in \mathcal{K}' \;\Big|\;
    \begin{aligned}
        &\mathrm{IoU}(b_{h_i}, \hat{b}_{h_g}) \geq 0.5 \\
        &\mathrm{IoU}(b_{o_i}, \hat{b}_{o_g}) \geq 0.5
    \end{aligned}
    \Big\}.
\end{equation}
The SR for the $g$-th ground-truth instance is then defined as: %
\begin{equation}
\text{SR}_g = \max_{(b_{h_i}, b_{o_i}, v'_i) \in \mathcal{K}'_g} \mathrm{sim}(v'_i, \hat{v}_g),
\end{equation}
where $\text{SR}_g = 0$ if $\mathcal{K}'_g = \emptyset$.
The overall SR is finally computed as the average across all ground-truth instances.

\subsection{Evaluation protocol}
\label{sec:metrics}
Inspired by evaluation protocols in other fields like multi-object tracking~\cite{meinhardt2022trackformer,Milan2016a}, we evaluate \task detection methods under two settings: the \emph{\publicsetting} and the \emph{\privatesetting} settings. In \textbf{annotated-box setting},  
the human and object bounding boxes are given as ground-truth, thereby isolating the \task prediction performance from the behavior of the detector.
Thus, the semantic association capability of the algorithm is evaluated independently from object localization errors. \textbf{Computed-box setting}
performs both human/object localization and interaction verb prediction in a unified manner.
This reflects the integral capabilities of the system in a practical deployment scenario.

\section{\method}
\label{sec:method}
To study \task we introduce AnyHOI, a training-free reference pipeline that leverages off-the-shelf Multimodal Large Language Models (MLLMs) to generate and extract human-object interactions from images.
We additionally propose to apply 
test-time to further enhance \lmms predictions, significantly boosting its performance with smaller \lmm models and enabling the detection of more subtle interactions.
As illustrated in \cref{fig:method}, we first detect all humans and objects in the image, assembling pairs of human-object combinations.
Then, we crop the image around each human-object pair and feed it to an \lmm along with a task-specific prompt, obtaining free-form descriptions of the scenes.
Finally, a post-generation pipeline processes the descriptions and extracts relationship triplets akin to the standard format of the HOI task.
This 
approach provides a reference point for studying the \task task and can be readily applied to a wide range of \lmms. \\ 
\vspace{-1em}

\noindent \textbf{Human-object pair detection.}
Given an RGB image $\mathcal{I}$, the first stage of \method aims to detect pairs of humans and objects depicted in it.
As it is common practice~\cite{Lei2023,Mao2023,Ning2023}, we rely on a frozen object detector $\text{f}_\text{D}: \mathcal{I} \to \mathcal{O}$ to obtain a set $\mathcal{O} = \{o_i\}_{i=1}^N$ of $N$ detections, where each detection $o_i = (b,~l)$ is defined by its bounding box $b \in \mathbb{R}^4$, and label $l$.
Subsequently, we obtain a set of human detections $\mathcal{O}_H = \{ o \in \mathcal{O} \mid l = \text{\texttt{"human"}} \}$ and construct 
interactions pairs $\mathcal{P} = \{ \langle o_h,~o_o\rangle ~ \mid ~ o_h \in \mathcal{O}_H, ~ o_o \in \mathcal{O}, ~ o_h \neq o_o \}$.
This comprehensive set of human-object 
candidate 
pairs will serve as the foundation for the interaction analysis of \method. \\ 
\vspace{-1em}

\noindent \textbf{Free-form generation.}
Let $\Psi(\mathcal{I}, b)$ represent the operation that crops the portion of image $\mathcal{I}$ delimited by bounding box $b$ (we test different visual prompts in~\cref{sec:ablations}), and let $f_{\text{\lmm}}: (\mathcal{I}, \mathcal{L}) \to \mathcal{L}'$ be the function implemented by the \lmm that maps visual $\mathcal{I}$ and textual $\mathcal{L}$ tokens to a sequence of output textual tokens $\mathcal{L}'$.
Given a human-object pair $p \in \mathcal{P}$, \method first crops the image to the union of their bounding boxes, \ie, $\mathcal{I}_p = \Psi(\mathcal{I}, b_h \cup b_o)$.
Then, the crop $\mathcal{I}_p$ and the textual prompt $Q_p = $~``\textit{What are the interactions between the person and the \texttt{obj}?}'', with \texttt{obj} being the label of the object in $p$, are fed to the \lmm to obtain an answer $\mathcal{A}_p = f_{\text{\lmm}}(\mathcal{I}_p, \mathcal{Q}_p)$.
Intuitively, $\mathcal{A}_p$ corresponds to a small paragraph describing 
the relationships between the person and the object in the prompt $\mathcal{Q}_p$ (we tested different textual prompts in~\cref{sec:ablations}). 
However, the generated answer does not follow any specific structure or template, and it may contain details and information that are not relevant for the HOI task (\eg, the color of the person's t-shirt).
To address this, \method implements a post-generation refinement that allows the extraction of HOI triplets from $\mathcal{A}_p$. 
\\ 
\vspace{-0.5em}

\noindent \textbf{Post-generation refinement.}
Given the free-form output of $f_{\text{\lmm}}$, a first naive approach to extract HOI triplets is to use a rule-based sentence parser such as spaCy~\cite{honnibal2020spacy} to extract verbs and objects from the generated output. However, we found those solutions to be inadequate to extract HOI triplets as they lack the flexibility to handle complex sentence structures.
To this end, \method employs FACTUAL~\cite{li2023factual}, an off-the-shelf text-to-graph Transformer-based model
$f_{\text{T2G}}: \mathcal{L} \to \mathcal{G}$ that converts free-form answers $\mathcal{A}_p$ into a set $\mathcal{G}$ of $\langle \text{\textit{subject}}, \text{\textit{verb}}, \text{\textit{object}} \rangle$ triplets.
Here, the subject is either a person or an object, and the verb represents a relationship between the subject and the object.
Due to the nature of $\mathcal{A}_p$, $f_{\text{T2G}}$ may also extract information unrelated to HOI, such as positional relationships between objects (\eg, $\langle \text{\textit{cup}}, \text{\textit{next to}}, \text{\textit{bottle}} \rangle$), or their attributes (\eg, $\langle \text{\textit{cup}}, \text{\textit{is}}, \text{\textit{white}} \rangle$).
To address this, \method applies a refinement step to prune the graph by retaining only triplets whose subjects are humans (\eg, \textit{man, woman, child}, etc.) and discarding those that contain copular or stative verbs (\eg, is, has), or whose objects do not match those from the original prompt $\mathcal{Q}_p$.
The remaining triplets form \method's final HOI predictions for the given human-object pair $p$.

\begin{figure}
\centering
\includegraphics[width=0.9\linewidth]{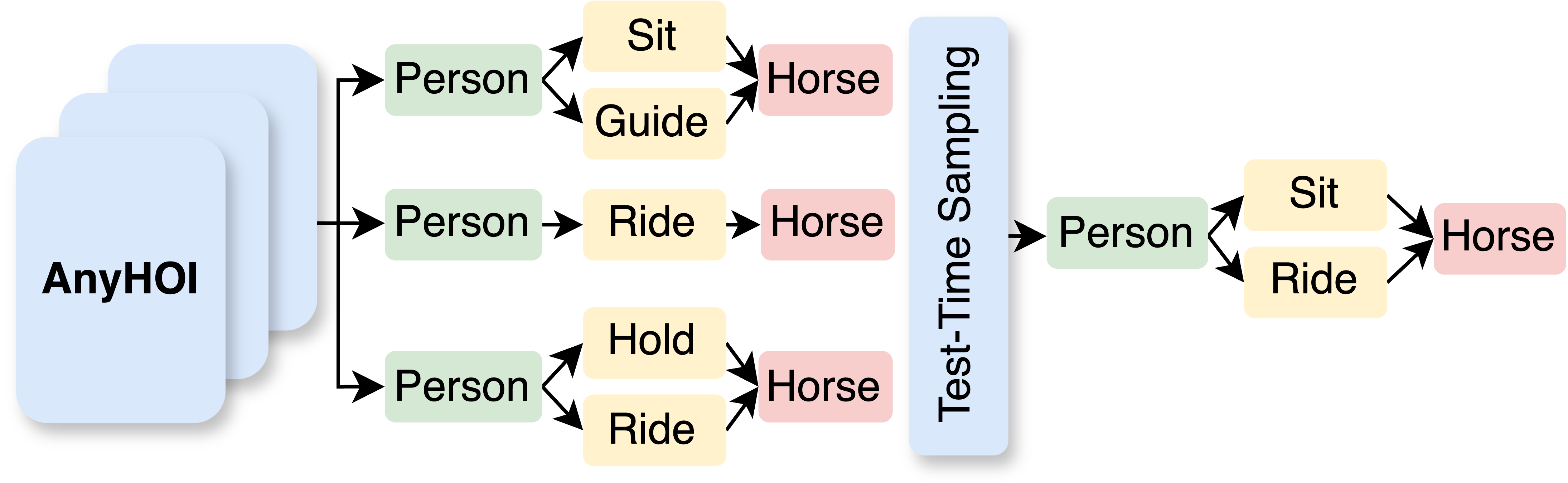}
\caption{
\textbf{The test-time compute strategy} generates multiple responses using its baseline \lmm, producing a diverse set of interaction proposals, which are aggregated and sampled based on their frequency, yielding the final predictions.
}
\label{fig:test_time}
\vspace{-1.5em}
\end{figure}

\vspace{-0.5em}
\noindent \textbf{Test-time compute.}
Current \lmms tend to primarily focus on the foreground elements of images~\cite{li2024llava,wang2024qwen2,huang2023visual,chen2024internvl}, often emphasizing a single interaction per generation $\mathcal{A}_p$.
As a consequence, this narrow focus overlooks subtler interactions that may be less prominent in the image.
Furthermore, this issue is exacerbated by the size of $f_{\text{\lmm}}$, with smaller models providing less informative outputs than their bigger alternatives.
Recently, test-time compute~\cite{snell2024scaling, muennighoff2025s1} has emerged as an effective approach to enhancing the reasoning capabilities of \lmms.
Building on this, we propose using test-time computing to improve the performance of \lmms and mitigate the limitations described above.
We follow the suggestions of the prior art on mathematical reasoning~\cite{hendrycks2021measuring} and visual understanding~\cite{liu2024mmbench} with \lmms and perform test-time compute by generating multiple samples for the same input, and then aggregating and analyzing the set of responses.
In particular, as shown in \cref{fig:test_time}, for a human-object pair $p$, we obtain a set of replies $\mathcal{A}_p^N$ by sampling $N$ responses from $f_{\text{\lmm}}$.
These generated answers are then subjected to post-generation filtering, yielding several sets of triplets.
To derive the final HOI interactions, we evaluate two approaches: keeping the $k$ most frequent triplets (referred to as \textit{Top-k} in Sec. \ref{sec:experiments}), and fitting a distribution on the triplets' frequency to later uniformly sample $k$ elements from it (referred to as \textit{Sampling} in Sec. \ref{sec:experiments}).

\section{Experiments}
\label{sec:experiments}
\noindent \textbf{Datasets.}
We evaluated \method on \hicodet and \vghoi.
\hicodet contains 117 verbs and 80 object classes, for a total of 600 different HOI triplets occurring in the dataset out of 9440 possible interactions, \ie, $\mathcal{M}$ has 600 non-zero entries.
For \hicodet, we follow the literature~\cite{zhang2022exploring,zhang2023exploring,Lei2023,lei2024exploring} and report the mAP on the 138 HOI categories with fewer than 10 training samples (Rare), the remaining 462 HOI categories (Non-rare), and all 600 categories (Full).
\vghoi is a subset of VG~\cite{krishna2017visual} 
for the HOI task. It contains 3,542 action classes and 5,385 object classes, for a total of 17,421 HOI triplets occurring in the dataset out of 19,073,670 possible interactions. 

\noindent \textbf{Baselines.}
Due to the unconstrained nature of the \task task, which violates core assumptions of traditional evaluation settings, some previous HOI methods cannot be directly applied and tested. Open-vocabulary models that rely on memory banks~\cite{Lei2023,tonini2025dynamic,lei2024exploring,lei2024ez} of verbs and objects cannot be tested with a vocabulary not known at training time.
Single-stage and two-stage models~\cite{Yang24,Liao2022,li2023neural,Ning2023,kim2025locality} require explicitly enumerating all possible combinations of verbs and objects, which is a major challenge for our setting, as it would result in tens of thousands of combinations for the \hicodet dataset and a staggering 19 million combinations for the \vghoi dataset. The sheer number of combinations required for these datasets makes their approach computationally infeasible.
Therefore, we adapted their classification head to use only verbs. 
On the other hand, DHD~\cite{Wu2024} features both an interaction classification head and a verb classification head, so we are able to swap the verb vocabulary at test time.
Additional details about how we adopted baselines for \taskacro are available in \cref{sec:additional_impl_det}.
When the \lmms are applied for the \task alone, we use spaCy~\cite{honnibal2020spacy} to perform POS tagging and extract nouns, verbs, and objects from the predicted answers, allowing us to extract HOI triplets from free-form generated text of \lmms.

\subsection{Implementation details}
\label{sec:impdet}

For the \privatesetting setting, we follow the literature~\cite{Lei2023,Zhang2022,zhang2023exploring} and adopt DETR~\cite{Zhang2022} for \hicodet, while for \vghoi we follow the approach of DHD~\cite{Wu2024} and use the open vocabulary detector GroundingDINO~\cite{liu2024grounding}. This ensures a fair comparison between our experiments and prior literature.
We filter detections with a confidence threshold below $0.2$ and sample a minimum of 3 and a maximum of 15 human and object instances. 
All experiments with \lmms use the same prompt defined in Sec.~\ref{sec:method} and we set the maximum output token length to 2048. During test-time computing, we set the generation temperature to $0.2$ to promote diversity and generate up to 64 predictions per pair. Then, we keep the top-$k$ triplets, with $k=10$.
For VLMs, we use CLIP ViT-L/14 and either the verb vocabulary of VG-HOI (3542 verbs) or WordNet (11026 verbs).
For WordNet, we removed verbs that are not valid for HOI (\eg, bordering), by asking an LLM~\cite{hurst2024gpt} whether a generic object could interact with the verb. In total, we kept 5004 verbs from WordNet.
Refer to the~\cref{sec:additional_impl_det} for the exact prompt used.
For $f_{\text{T2G}}$, we use FACTUAL~\cite{li2023factual}, a state-of-the-art method for text-to-scene-graph.
We evaluate the test-time performance on the LLaVA OV family, focusing specifically on the 0.5B model version. Our goal is to explore whether smaller models can match or even outperform their larger counterparts, achieving comparable results with only a fraction of the parameters.
For all settings, we evaluated mAP with thresholds $\mathcal{T} = \{0.6, 0.7, 0.8, 0.9, 0.95\}$.

\begin{table}[]
\caption{\task performance on VG-HOI~\cite{Wu2024} dataset under both annotated and computed-box settings.}
\vspace{-1em}
\centering
\Large
\resizebox{0.9\columnwidth}{!}{%
\begin{tabular}{lcl@{\hspace{5pt}}l|l@{\hspace{4pt}}l}
\toprule
\multicolumn{2}{c|}{\multirow{2}{*}{\textbf{Method}}}                                        & \multicolumn{2}{c|}{\textbf{Annotated-box}}                                                   & \multicolumn{2}{c}{\textbf{Computed-box}}                                                     \\
\multicolumn{2}{c|}{}                                                                        & \textbf{mAP (\%) $\uparrow$} & \textbf{SR (x100) $\uparrow$} & \textbf{mAP (\%) $\uparrow$} & \textbf{SR (x100) $\uparrow$} \\ \midrule
\rowcolor{ourtablecolor} \multicolumn{2}{l|}{Random}                                                                 & 0.30                          & 35.89                                                   & 0.05                          & 6.68                                                    \\
\multicolumn{1}{c}{}                                         & \multicolumn{1}{c|}{VG}  & 9.80                          & 55.83                                                   & 1.45                          & 9.71                                                   \\
\multicolumn{1}{l}{\multirow{-2}{*}{CLIP~\cite{Radford2021}}} & \multicolumn{1}{c|}{WN} & 2.95                          & 39.75                                                   & 0.37                          & 7.22                                                    \\
\rowcolor{ourtablecolor} \multicolumn{1}{c}{}                                         & \multicolumn{1}{c|}{VG}  & 8.73                          & 52.85                                                   & 1.06                          & 9.89                                                    \\
\rowcolor{ourtablecolor} \multicolumn{1}{c}{\multirow{-2}{*}{DHD~\cite{Wu2024}}}                                         & \multicolumn{1}{c|}{WN} & 8.86                          & 50.36                                                   & 1.98                          & 10.31                                                  \\
\multicolumn{2}{l|}{CogVLM2 19B~\cite{hong2024cogvlm2}} & 2.28 & 7.53 & 0.22 & 1.23 \\
\multicolumn{2}{l|}{\textbf{+ \method (Ours)}} & 18.20 \gooddelta{15.92} & 60.25 \gooddelta{52.72} & 2.11 \gooddelta{1.89} & 10.98 \gooddelta{9.75} \\
\rowcolor{ourtablecolor} \multicolumn{2}{l|}{Idefics2 8B~\cite{laurenccon2025matters}} & 12.54 & 39.96 & 1.13 & 6.68 \\
\rowcolor{ourtablecolor} \multicolumn{2}{l|}{\textbf{+ \method (Ours)}} & 19.60 \gooddelta{7.06} & 59.65 \gooddelta{19.69} & 1.90 \gooddelta{0.77} & 10.43 \gooddelta{3.75} \\
\multicolumn{2}{l|}{InternVL2 2B~\cite{chen2024internvl}} & 6.29 & 25.3 & 1.07 & 5.26 \\
\multicolumn{2}{l|}{\textbf{+ \method (Ours)}} & 15.34 \gooddelta{9.05} & 50.50 \gooddelta{25.20} & 2.43 \gooddelta{1.36} & 9.71 \gooddelta{4.45} \\
\rowcolor{ourtablecolor} \multicolumn{2}{l|}{InternVL2 4B~\cite{chen2024internvl}} & 7.87 & 33.22 & 0.77 & 5.2 \\
\rowcolor{ourtablecolor} \multicolumn{2}{l|}{\textbf{+ \method (Ours)}} & 17.60 \gooddelta{9.73} & 58.73 \gooddelta{25.51} & 2.25 \gooddelta{1.48} & 10.83 \gooddelta{5.63} \\
\multicolumn{2}{l|}{InternVL2 8B~\cite{chen2024internvl}} & 10.36 & 39.14 & 1.03 & 6.61 \\
\multicolumn{2}{l|}{\textbf{+ \method (Ours)}} & 19.83 \gooddelta{9.47} & 62.37 \gooddelta{23.23} & 2.78 \gooddelta{1.75} & 11.76 \gooddelta{5.15} \\
\rowcolor{ourtablecolor} \multicolumn{2}{l|}{InstructBLIP 7B~\cite{huang2023visual}} & 2.37 & 9.03 & 0.38 & 1.59 \\
\rowcolor{ourtablecolor} \multicolumn{2}{l|}{\textbf{+ \method (Ours)}} & 15.71 \gooddelta{13.34} & 56.37 \gooddelta{47.34} & 2.61 \gooddelta{2.23} & 10.90 \gooddelta{9.31} \\
\multicolumn{2}{l|}{InstructBLIP 13B~\cite{huang2023visual}} & 1.87 & 8.54 & 0.33 & 1.6 \\
\multicolumn{2}{l|}{\textbf{+ \method (Ours)}} & 15.20 \gooddelta{13.33} & 57.14 \gooddelta{48.60} & 2.44 \gooddelta{2.11} & 10.86 \gooddelta{9.26} \\
\rowcolor{ourtablecolor} \multicolumn{2}{l|}{Phi3V 4B~\cite{abdin2024phi}} & 1.78 & 5.57 & 0.22 & 0.75 \\
\rowcolor{ourtablecolor} \multicolumn{2}{l|}{\textbf{+ \method (Ours)}} & 17.37 \gooddelta{15.59} & 54.35 \gooddelta{48.78} & 1.87 \gooddelta{1.65} & 9.58 \gooddelta{8.83} \\
\multicolumn{2}{l|}{Qwen2-VL 2B~\cite{wang2024qwen2}} & 4.56 & 17.68 & 0.55 & 2.71 \\
\multicolumn{2}{l|}{\textbf{+ \method (Ours)}} & 14.16 \gooddelta{9.60} & 47.64 \gooddelta{29.96} & 2.15 \gooddelta{1.60} & 9.94 \gooddelta{7.23} \\
\rowcolor{ourtablecolor} \multicolumn{2}{l|}{Qwen2-VL 7B~\cite{wang2024qwen2}} & 9.46 & 38.32 & 0.87 & 6.07 \\
\rowcolor{ourtablecolor} \multicolumn{2}{l|}{\textbf{+ \method (Ours)}} & 19.61 \gooddelta{10.15} & 63.76 \gooddelta{25.44} & 2.76 \gooddelta{1.89} & 12.14 \gooddelta{6.07} \\
\multicolumn{2}{l|}{LLaVA OV 0.5B~\cite{li2024llava}} & 1.21 & 4.41 & 0.15 & 0.71 \\
\multicolumn{2}{l|}{\textbf{+ \method (Ours)}} & 12.76 \gooddelta{11.55} & 42.70 \gooddelta{38.29} & 1.99 \gooddelta{1.84} & 8.10 \gooddelta{7.39} \\
\rowcolor{ourtablecolor} \multicolumn{2}{l|}{LLaVA OV 7B~\cite{li2024llava}} & 10.54 & 33.53 & 0.93 & 5.97 \\
\rowcolor{ourtablecolor} \multicolumn{2}{l|}{\textbf{+ \method (Ours)}} & 21.65 \gooddelta{11.11} & 64.91 \gooddelta{31.38} & 2.64 \gooddelta{1.71} & 12.12 \gooddelta{6.15} \\
\bottomrule
\multicolumn{6}{c}{``VG'' and ``WN'' stand for VG-HOI and WordNet, respectively.}
\end{tabular}%
}
\label{tab:main_vghoi_new}
\vspace{-1.5em}
\end{table}

\subsection{Benchmarking MLLMs for \task}
\label{sec:experiments-benchmarking}
\noindent \textbf{Results on VG-HOI}.
We compare the HOI SOTA method DHD~\cite{Wu2024}, CLIP~\cite{Radford2021}, and various \lmms~\cite{hong2024cogvlm2, laurenccon2025matters, chen2024internvl, huang2023visual, abdin2024phi, wang2024qwen2, li2024llava} on \vghoi in~\cref{tab:main_vghoi_new} for the \publicsetting and 
\privatesetting settings of \task.
We also test the performance of the aforementioned \lmms when incorporating \method, presenting a training-free baseline for solving \task.
Finally, recall that we use the VG-HOI and WordNet vocabularies for~\cite{Radford2021} and~\cite{Wu2024} to be able to adopt them for \task. 

Our findings show that off-the-shelf \lmms can achieve results comparable to state-of-the-art (SOTA) HOI models such as DHD on the \task task (\eg, $12.54/39.96$ vs. $8.86/50.36$ mAP/SR for Idefics2 8B and DHD, respectively).
However, these results remain relatively low and are only moderately better than random performance (\eg, a gap of $+8.56/+14.14$ mAP/SR for DHD), highlighting the inherent difficulty of the task and the impact of removing the vocabulary prior.

Injecting \method into the \lmms leads to substantial improvements, with performance gains reaching up to $15.92/52.72$ mAP/SR.
With \method, models such as Qwen2-VL 7B, LLaVA OV 7B, and InternVL2 8B achieve competitive results with DHD and CLIP across both the \privatesetting and \publicsetting settings.
Notably, even the smallest \lmm (LLaVA OV 0.5B~\cite{li2024llava}) achieves higher mAP than both CLIP and DHD in both settings when augmented with \method.
Finally, although the \publicsetting setting is more challenging due to noisy detections/pairs and a greater susceptibility to false positives, both \lmms and \lmms+\method remain competitive with existing HOI baselines under the \task setting.

\begin{table}[!ht]
\caption{\task performance on HICO-DET~\cite{Chao2018} dataset under \privatesetting setting.}
\vspace{-1em}
\centering
\Large
\resizebox{0.9\linewidth}{!}{
\begin{tabular}{lc|l@{\hspace{5pt}}l@{\hspace{5pt}}l@{\hspace{5pt}}|l}
\toprule
\multicolumn{2}{c|}{\multirow{2}{*}{\textbf{Method}}}  & \multicolumn{3}{c|}{\textbf{mAP (\%)} $\uparrow$} & \multicolumn{1}{c}{\textbf{SR} $\mathbf{\uparrow}$}  \\
&& \multicolumn{1}{c}{Full} & \multicolumn{1}{c}{Rare} & \multicolumn{1}{c|}{Non-rare} & \multicolumn{1}{c}{\textbf{(x100)}} \\
\midrule
\multicolumn{2}{l|}{Random} & 0.69 & 0.27 & 0.81 & 22.22  \\
\rowcolor{ourtablecolor} & VG & 3.44 & 7.42 & 2.26 & 26.46   \\
\rowcolor{ourtablecolor} \multirow{-2}{*}{{CLIP~\cite{Radford2021}}} & WN & 3.13 & 7.90 & 1.70 & 23.57  \\
\multirow{2}{*}{{GEN-VLKT~\cite{Liao2022}}} & VG & 4.99 & 5.18 & 4.93 & 29.77 \\
& WN & 3.67 & 2.09 & 4.14 & 28.89  \\
\rowcolor{ourtablecolor} & VG & 2.76 & 2.27 & 2.90 & 26.86  \\
\rowcolor{ourtablecolor} \multirow{-2}{*}{{HOICLIP~\cite{Ning2023}}} & WN & 1.52 & 2.15 & 1.34 & 21.21  \\
& VG & 2.16 & 0.87 & 2.55 & 20.96  \\
\multirow{-2}{*}{{LogicHOI~\cite{li2023neural}}} & WN & 1.49 & 0.47 & 1.79 & 19.23 \\
\rowcolor{ourtablecolor} & VG & 6.75 & 11.28 & 5.40 & 29.85  \\ 
\rowcolor{ourtablecolor} \multirow{-2}{*}{{MP-HOI~\cite{Yang24}}} & WN & 5.47 & 8.25 & 4.64 & 27.71  \\
& VG & 6.97 & 2.72 & 8.24 & 30.92  \\
\multirow{-2}{*}{{LAIN~\cite{kim2025locality}}} & WN & 1.05 & 1.05 & 1.05 & 22.83  \\
\rowcolor{ourtablecolor} \multicolumn{2}{l|}{CogVLM2~\cite{hong2024cogvlm2}} & 1.34 & 0.67 & 1.54 & 7.12 \\
\rowcolor{ourtablecolor} \multicolumn{2}{l|}{\textbf{+ \method (Ours)}} & 8.10 \gooddelta{6.76} & 7.43 \gooddelta{6.76} & 8.31 \gooddelta{6.77} & 29.17 \gooddelta{22.05} \\
\multicolumn{2}{l|}{InternVL2 8B~\cite{chen2024internvl}} & 1.91 & 1.2 & 2.12 & 8.06 \\
\multicolumn{2}{l|}{\textbf{+ \method (Ours)}} & 6.47 \gooddelta{4.56} & 5.79 \gooddelta{4.59} & 6.68 \gooddelta{4.56} & 26.84 \gooddelta{18.78} \\
\rowcolor{ourtablecolor} \multicolumn{2}{l|}{Qwen2-VL 7B~\cite{wang2024qwen2}} & 4.03 & 4.54 & 3.87 & 11.65 \\
\rowcolor{ourtablecolor} \multicolumn{2}{l|}{\textbf{+ \method (Ours)}} & 9.35 \gooddelta{5.32} & 10.62 \gooddelta{6.08} & 8.97 \gooddelta{5.10} & 32.92 \gooddelta{21.27} \\
 \multicolumn{2}{l|}{LLaVA OV 0.5B~\cite{li2024llava}} & 1.32 & 0.19 & 1.66 & 5.73 \\
\multicolumn{2}{l|}{\textbf{+ \method (Ours)}} & 6.43 \gooddelta{5.11} & 5.09 \gooddelta{4.90} & 6.84 \gooddelta{5.18} & 20.04 \gooddelta{14.31} \\
\rowcolor{ourtablecolor}\multicolumn{2}{l|}{LLaVA OV 7B~\cite{li2024llava}} & 2.72 & 2.49 & 2.79 & 8.05 \\
\rowcolor{ourtablecolor}\multicolumn{2}{l|}{\textbf{+ \method (Ours)}} & 8.73 \gooddelta{6.01} & 10.56 \gooddelta{8.07} & 8.19 \gooddelta{5.40} & 28.59 \gooddelta{20.54} \\
\bottomrule
\multicolumn{6}{c}{``VG'' and ``WN'' stand for VG-HOI and WordNet, respectively.}
\end{tabular}%
}
\label{tab:main_hicodet_detection}
\end{table}

\begin{table}[!ht]
\caption{\task performance on HICO-DET~\cite{Chao2018} dataset under \publicsetting setting.}
\vspace{-1em}
\Large
\centering
\resizebox{0.9\linewidth}{!}{
\begin{tabular}{cc|l@{\hspace{5pt}}l@{\hspace{5pt}}l|l}
\toprule
\multicolumn{2}{c|}{\multirow{2}{*}{\textbf{Method}}}  & \multicolumn{3}{c|}{\textbf{mAP (\%)} $\uparrow$} &  \multicolumn{1}{c}{\textbf{SR} $\mathbf{\uparrow}$}   \\
&& \multicolumn{1}{c}{Full} & \multicolumn{1}{c}{Rare} & \multicolumn{1}{c|}{Non-rare} &  \multicolumn{1}{c}{\textbf{(x100)}}  \\
\midrule
\multicolumn{2}{l|}{Random} & 0.79 & 0.27 & 0.95 & 31.68 \\
\rowcolor{ourtablecolor} & VG & 5.49 & 11.47 & 3.44 & 37.55 \\
\rowcolor{ourtablecolor} \multirow{-2}{*}{{CLIP~\cite{Radford2021}}} & WN & 4.51 & 11.25 & 2.50 & 33.86 \\
\multicolumn{2}{l|}{CogVLM2 19B~\cite{hong2024cogvlm2}} & 2.08 & 1.57 & 2.23 & 9.94 \\
\multicolumn{2}{l|}{\textbf{\textbf{+ \method (Ours)}}} & 13.01 \gooddelta{10.93} & 12.45 \gooddelta{10.88} & 13.18 \gooddelta{10.95} & 40.96 \gooddelta{31.02} \\
\rowcolor{ourtablecolor} \multicolumn{2}{l|}{Idefics2 8B~\cite{laurenccon2025matters}} & 6.59 & 5.97 & 6.77 & 20.29 \\
\rowcolor{ourtablecolor} \multicolumn{2}{l|}{\textbf{\textbf{+ \method (Ours)}}} & 10.28 \gooddelta{3.69} & 10.15 \gooddelta{4.18} & 10.32 \gooddelta{3.55} & 34.27 \gooddelta{13.98} \\
\multicolumn{2}{l|}{InternVL 2B~\cite{chen2024internvl}} & 2.91 & 1.68 & 3.28 & 11.41 \\
\multicolumn{2}{l|}{\textbf{\textbf{+ \method (Ours)}}} & 10.21 \gooddelta{7.30} & 8.87 \gooddelta{7.19} & 10.61 \gooddelta{7.33} & 36.59 \gooddelta{25.18} \\
\rowcolor{ourtablecolor} \multicolumn{2}{l|}{InternVL 4B~\cite{chen2024internvl}} & 3.43 & 1.71 & 3.94 & 11.28 \\
\rowcolor{ourtablecolor} \multicolumn{2}{l|}{\textbf{\textbf{+ \method (Ours)}}} & 9.96 \gooddelta{6.53} & 8.35 \gooddelta{6.64} & 10.45 \gooddelta{6.51} & 36.85 \gooddelta{25.57} \\
\multicolumn{2}{l|}{InternVL 8B~\cite{chen2024internvl}} & 2.88 & 1.33 & 3.34 & 11.29 \\
\multicolumn{2}{l|}{\textbf{\textbf{+ \method (Ours)}}} & 11.25 \gooddelta{8.37} & 10.68 \gooddelta{9.35} & 11.42 \gooddelta{8.08} & 37.84 \gooddelta{26.55} \\
\rowcolor{ourtablecolor} \multicolumn{2}{l|}{InstructBLIP 7B~\cite{huang2023visual}} & 4.2 & 3.95 & 4.28 & 10.68 \\
\rowcolor{ourtablecolor} \multicolumn{2}{l|}{\textbf{\textbf{+ \method (Ours)}}} & 11.53 \gooddelta{7.33} & 13.02 \gooddelta{9.07} & 11.09 \gooddelta{6.81} & 40.52 \gooddelta{29.84} \\
\multicolumn{2}{l|}{InstructBLIP 13B~\cite{huang2023visual}} & 1.91 & 0.37 & 2.38 & 8.04 \\
\multicolumn{2}{l|}{\textbf{\textbf{+ \method (Ours)}}} & 11.09 \gooddelta{9.18} & 12.93 \gooddelta{12.56} & 10.54 \gooddelta{8.16} & 43.57 \gooddelta{35.53} \\
\rowcolor{ourtablecolor} \multicolumn{2}{l|}{Phi3V 4B~\cite{abdin2024phi}} & 2.07 & 1.22 & 2.32 & 8.77 \\
\rowcolor{ourtablecolor} \multicolumn{2}{l|}{\textbf{\textbf{+ \method (Ours)}}} & 10.78 \gooddelta{8.71} & 10.64 \gooddelta{9.42} & 10.82 \gooddelta{8.50} & 39.99 \gooddelta{31.22} \\
\multicolumn{2}{l|}{Qwen2-VL 2B~\cite{wang2024qwen2}} & 3.35 & 2.21 & 3.69 & 11.08 \\
\multicolumn{2}{l|}{\textbf{\textbf{+ \method (Ours)}}} & 12.89 \gooddelta{9.54} & 13.45 \gooddelta{11.24} & 12.72 \gooddelta{9.03} & 41.21 \gooddelta{30.13} \\
\rowcolor{ourtablecolor} \multicolumn{2}{l|}{Qwen2-VL 7B~\cite{wang2024qwen2}} & 6.48 & 7.94 & 6.05 & 17.89 \\
\rowcolor{ourtablecolor} \multicolumn{2}{l|}{\textbf{\textbf{+ \method (Ours)}}} & 14.68 \gooddelta{8.20} & 16.92 \gooddelta{8.98} & 14.02 \gooddelta{7.97} & 45.74 \gooddelta{27.85} \\
\multicolumn{2}{l|}{LLaVA OV 0.5B~\cite{li2024llava}} & 1.32 & 0.14 & 1.67 & 7.53 \\
 \multicolumn{2}{l|}{\textbf{\textbf{+ \method (Ours)}}} & 13.46 \gooddelta{12.14} & 15.12 \gooddelta{14.98} & 12.96 \gooddelta{11.29} & 39.91 \gooddelta{32.38} \\
\rowcolor{ourtablecolor} \multicolumn{2}{l|}{LLaVA OV 7B~\cite{li2024llava}} & 4.25 & 4.85 & 4.07 & 11.7 \\
\rowcolor{ourtablecolor} \multicolumn{2}{l|}{\textbf{\textbf{+ \method (Ours)}}} & 13.01 \gooddelta{8.76} & 14.12 \gooddelta{9.27} & 12.68 \gooddelta{8.61} & 39.89 \gooddelta{28.19} \\
\bottomrule
\multicolumn{6}{c}{``VG'' and ``WN'' stand for VG-HOI and WordNet, respectively.}
\end{tabular}%
}
\vspace{-0.5em}
\label{tab:main_hicodet_classification}
\vspace{-0.5em}
\end{table}

\noindent \textbf{Results on HICO-DET}. 
\cref{tab:main_hicodet_classification,tab:main_hicodet_detection} present the results for HOI SOTA methods~\cite{Yang24,Liao2022,li2023neural,kim2025locality,Ning2023}, CLIP~\cite{Radford2021}, and various \lmms~\cite{hong2024cogvlm2, laurenccon2025matters, chen2024internvl, huang2023visual, abdin2024phi, wang2024qwen2, li2024llava} with and without \method
on the HICO-DET dataset for the \publicsetting and \privatesetting settings, respectively.
For the \privatesetting setting (\cref{tab:main_hicodet_detection}), we focus our evaluation on \lmms that demonstrate superior performance in both the \publicsetting setting.

Consistent with our findings on VG-HOI, \lmms serve as a strong alternative to traditional HOI models on the \task task. 
Their advantage becomes especially pronounced when combined with \method, which consistently enhances performance across all metrics and both settings, achieving results competitive with standard HOI approaches such as CLIP~\cite{Radford2021}, GEN-VLKT~\cite{Liao2022}, HOICLIP~\cite{Ning2023}, LogicHOI~\cite{li2023neural}, MP-HOI~\cite{Yang24}, and LAIN~\cite{kim2025locality}.

\subsection{Test-time \lmms for \task}
\label{sec:experiments-test_time}
We evaluate the test-time computing component of \method on \hicodet (\cref{tab:main_test_time_hicodet}) and \vghoi (\cref{tab:main_test_time_vghoi}), under the \publicsetting and \privatesetting settings using LLaVA OV 0.5B as our baseline.
As shown, it brings a significant boost to all metrics, with a +11.70\% improvement in mAP and a +30.12 increase in $SR$ on \vghoi w.r.t. \method without test-time, and +7.29\% improvement in mAP and a +23.63\% $SR$ improvement on \hicodet.
We emphasize that test-time compute serves as an analysis tool for probing \lmm capabilities under \task, not as a deployment requirement.
\begin{table}[!t]
\caption{Performance of \method with test-time using LLaVA OV 0.5B on \hicodet dataset.}
\centering
\resizebox{0.7\columnwidth}{!}{%
\begin{tabular}{@{}cl|l@{\hspace{5pt}}l@{\hspace{5pt}}l|l@{}}
\toprule
\multicolumn{2}{c|}{\multirow{2}{*}{\textbf{Method}}}  & \multicolumn{3}{c|}{\textbf{mAP Avg. (\%)} $\uparrow$} &  \multicolumn{1}{c}{\textbf{SR} $\mathbf{\uparrow}$}  \\
& & \multicolumn{1}{c}{Full} & \multicolumn{1}{c}{Rare} & \multicolumn{1}{c|}{Non-rare} &  \multicolumn{1}{c}{\textbf{(x100)}} \\
\midrule
\multicolumn{1}{l}{} & CLIP~\cite{Radford2021} (VG) & 5.49 & 11.47 & 3.44 & 37.55 \\
\multicolumn{1}{l}{} & \method & 13.46 & 15.12 & 12.96 & 39.91 \\
\multicolumn{1}{c}{\multirow{-3}{*}{\rotatebox[origin=c]{90}{Annot.}}} & + TT & 20.75 \gooddelta{7.29} & 25.25 \gooddelta{10.13} & 19.40 \gooddelta{6.44} & 63.54 \gooddelta{23.63} \\

\midrule

\rowcolor{ourtablecolor} \multicolumn{1}{l}{} & LAIN$_\text{CVPR25}$~\cite{kim2025locality} (VG) & 6.97 & 2.72 & 8.24 & 30.92 \\
\rowcolor{ourtablecolor}  \multicolumn{1}{l}{} & \method & 6.43 & 5.09 & 6.84 & 20.04 \\
\rowcolor{ourtablecolor}  \multicolumn{1}{c}{\multirow{-3}{*}{\rotatebox[origin=c]{90}{Comp.}}} & + TT & 10.30 \gooddelta{3.87} & 11.35 \gooddelta{6.26} & 9.99 \gooddelta{3.15} & 36.25 \gooddelta{16.21} \\
\bottomrule
\multicolumn{6}{c}{``Annot.'' and ``Comp.'' stand for \publicsetting and \privatesetting respectively.}
\end{tabular}%
}
\vspace{-0.5em}
\label{tab:main_test_time_hicodet}
\end{table}

\begin{table}[!t]
\caption{Performance of \method with test-time using LLaVA OV 0.5B on \vghoi dataset.}
\centering
\resizebox{0.7\columnwidth}{!}{%
\begin{tabular}{@{}cl|l|l@{}}
\toprule
\multicolumn{2}{c|}{\textbf{Method}}  & \multicolumn{1}{c|}{\textbf{mAP Avg. (\%)} $\uparrow$} & \textbf{SR (x100)} $\mathbf{\uparrow}$  \\
\midrule
\multicolumn{1}{l}{} & DHD$_\text{AAAI24}$~\cite{Wu2024} (WN) & 8.86 & 50.36 \\
\multicolumn{1}{l}{} & \method & 12.76 & 42.70 \\
\multicolumn{1}{c}{\multirow{-3}{*}{\rotatebox[origin=c]{90}{Annot.}}} & + TT & 24.46 \gooddelta{11.70} & 72.82 \gooddelta{30.12} \\

\midrule

\rowcolor{ourtablecolor} \multicolumn{1}{l}{} & DHD$_\text{AAAI24}$~\cite{Wu2024} (WN) & 1.98 & 10.31 \\
\rowcolor{ourtablecolor}  \multicolumn{1}{l}{} & \method & 1.99 & 8.10 \\
\rowcolor{ourtablecolor}  \multicolumn{1}{c}{\multirow{-3}{*}{\rotatebox[origin=c]{90}{Comp.}}} & + TT & 3.47 \gooddelta{1.48} & 12.99 \gooddelta{4.89} \\
\bottomrule
\multicolumn{4}{l}{``Annot.'' and ``Comp.'' stand for \publicsetting and \privatesetting respectively.}
\end{tabular}%
}
\vspace{-0.5em}
\label{tab:main_test_time_vghoi}
\end{table}

\noindent \textbf{Sampling vs. Top-k.}
Recall that in Sec. \ref{sec:method} we introduced two strategies: \textit{Top-k} and \textit{Sampling}, for determining the final HOI triplet prediction when our test-time strategy is applied. 
\begin{table}[!t]
\caption{Performance of \method + TT using different test-time sampling strategies on HICO-DET~\cite{Chao2018} dataset under \publicsetting setting.}
\centering
\resizebox{0.7\columnwidth}{!}{%
\begin{tabular}{ccl|ccc|c}
\toprule
\multicolumn{2}{c|}{\multirow{2}{*}{\textbf{Method}}} & \multicolumn{1}{c|}{\multirow{2}{*}{\textbf{Method}}}  & \multicolumn{3}{c|}{\textbf{mAP Avg. (\%)} $\uparrow$} & \textbf{SR} $\mathbf{\uparrow}$   \\

\multicolumn{2}{c|}{} & {} & Full & Rare & Non-rare & {\textbf{(x100)}} \\

\midrule

{} & \multicolumn{1}{l}{\multirow{2}{*}{0.5B~\cite{li2024llava}}} & Sampling & 18.90 & 21.71 & 18.06 & 57.31 \\

{} & {}                       & Top-k  & 20.75 & 25.25 & 19.40 & 63.54  \\

\rowcolor{ourtablecolor} \cellcolor{white}{} & {} & Sampling & 21.32 & 26.86 & 19.67 & 56.46 \\

\rowcolor{ourtablecolor} {\multirow{-4}{*}{\rotatebox[origin=c]{90}{\cellcolor{white}LLaVA}}} & \multicolumn{1}{c}{\multirow{-2}{*}{7B~\cite{li2024llava}}}                       & Top-k  & \textbf{23.01} & \textbf{29.03} &  \textbf{21.20} & 61.14  \\

\bottomrule
\end{tabular}%
}
\label{tab:abl_test_time_strategy}
\vspace{-2em}
\end{table}

\cref{tab:abl_test_time_strategy} compares these two strategies for LLaVA 0.5B, the smallest \lmm among those we used, and for comparison purposes, LLaVA OV 7B, which belongs to the same model family on \hicodet. 
For both methods, Top-k performs better in all metrics, demonstrating that prioritizing the most frequent responses of \lmms leads to improved performance.

\noindent \textbf{The impact of $\pmb{k}$.}
We analyzed the impact of $k$ (\ie, the number of interaction triplets considered at test time) on the performance of CLIP~\cite{Radford2021} and LLaVA OV 0.5B~\cite{li2024llava} + \method, evaluating them with $k$ set to 1, 2, 5, and 10.
The results are shown in~\cref{tab:abl_k} of \supmat, and include all proposed metrics on the HICO-DET~\cite{Chao2018} dataset and the \publicsetting setting. Increasing $k$ generally leads to improved performance on mAP and $SR$, indicating that incorporating multiple candidate triplets enhances flexibility w.r.t. vocabulary and recall.
While CLIP benefits from higher $k$ values, its overall performance remains relatively low compared to our approach. Using WordNet instead of \vghoi results in consistently lower scores across all settings. LLaVA OV 0.5B achieves higher scores than CLIP-based methods, reaching 20.75\% mAP Avg. at $k=10$. It also shows improved detection of rare interactions, peaking at 33.11\% with $k=2$.

\noindent \textbf{Increasing the number of generations at test-time.}
As shown in~\cref{fig:abl_test_time} of \supmat, \method + TT continues to improve as the number of sampled answers increases during test time. 
With only 2 generations per sample, LLaVA 0.5B results already improving by +2\% mAP, and +6\% with 4 generations, over \method without test-time. This demonstrates that test-time sampling with a temperature of 0.2 enables exploration of different generation paths in the \lmms, potentially capturing multiple distinct interactions.

\begin{figure*}[!ht]
    \centering
    \includegraphics[width=0.87\linewidth]{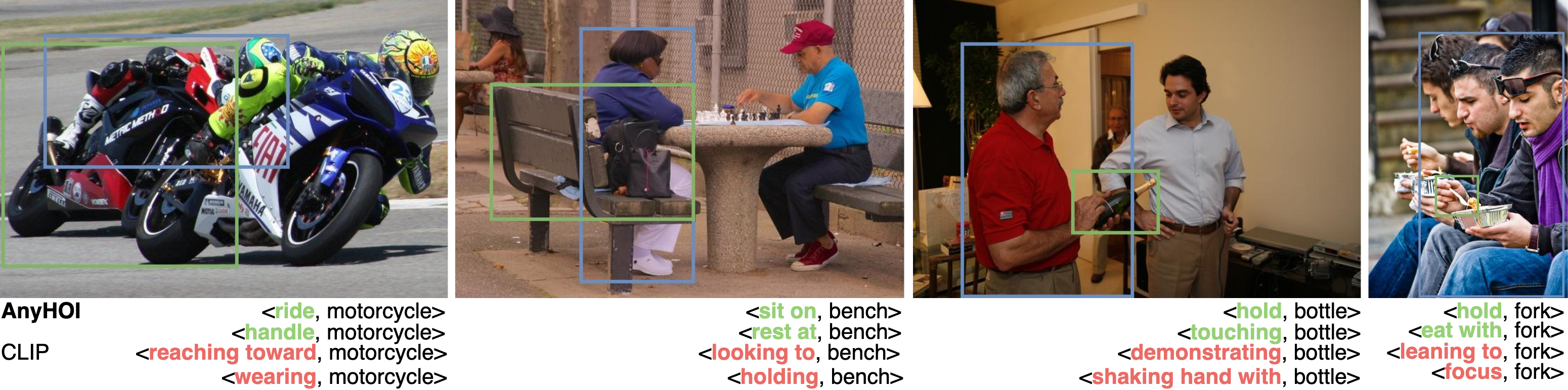}
    \caption{Qualitative results of \method and CLIP~\cite{Radford2021} on the \hicodet dataset}
    \label{fig:main_qualitatives}
    \vspace{-1.5em}
\end{figure*}

\subsection{Ablation studies}
\label{sec:ablations}
\noindent \textbf{On different visual prompting strategies}.
To examine the impact of the vision modality in \task, we conduct a study testing various visual augmentation methods (red circle, reverse blur, crop w/ mask, and crop). For \textit{red circle}~\cite{Shtedritski23circle}, we circle the area in which the human and the object are to elicit model reasoning on the area of interest. We follow the same intuition for \textit{reverse blur} (\ie we blur all image content outside of human and object), and \textit{crop} (\ie we crop the image around the bounding box that includes both human and object). For \textit{crop with masking}, we mask all pixels outside of the human and object boxes to remove any potential distractors and/or objects that may appear inside the union box.
Lastly, we include a ``\textit{blind}'' experiment, where we remove the visual signal from the \lmms by replacing the input image $\mathcal{I}_p$ with a black image.
\begin{table}[!t]
\caption{Evaluation of visual prompting techniques using LLaVA OV 0.5B on \hicodet dataset under \publicsetting setting.}
\centering
\resizebox{0.6\columnwidth}{!}{%
\begin{tabular}{@{}l|ccc|c@{}}
\toprule
\multicolumn{1}{c|}{\textbf{Vision}} & \multicolumn{3}{c|}{\textbf{mAP Avg. (\%)} $\uparrow$} &  \multicolumn{1}{c}{\textbf{SR} $\mathbf{\uparrow}$} \\
\multicolumn{1}{c|}{\textbf{Prompts}} & Full & Rare & Non-rare &  \multicolumn{1}{c}{\textbf{(x100)}} \\ \midrule
\rowcolor{ourtablecolor} Blind & 4.12 & 0.53 & 5.19 & 37.25 \\
Red circle  & 12.44 & 13.76 & 12.04 & 38.25 \\
\rowcolor{ourtablecolor} Reverse blur  & 13.06 & 13.55 & 12.92 & 38.33 \\
Crop w/ mask  & 13.01 & 13.90 & 12.74 & 38.21 \\
\rowcolor{ourtablecolor} Crop  & \textbf{13.46} & \textbf{15.12} & \textbf{12.96} & \textbf{39.91} \\
\bottomrule
\multicolumn{5}{c}{``Blind'' refers to replacing the input with a black image.}
\end{tabular}%
}
\label{tab:abl_visual_augmentation_lmm}
\end{table}

\begin{table}[!t]
\caption{Performance of \method using different prompts on HICO-DET~\cite{Chao2018} dataset under \publicsetting setting.}
\centering
\resizebox{0.6\columnwidth}{!}{%
\begin{tabular}{lc|ccc|c}
\toprule
\multicolumn{1}{c}{\textbf{Method}} & \textbf{PP} & \multicolumn{3}{c|}{\textbf{mAP Avg. (\%)} $\uparrow$} & \textbf{SR} $\mathbf{\uparrow}$  \\
& & \multicolumn{1}{c}{Full} & \multicolumn{1}{c}{Rare} & \multicolumn{1}{c|}{Non-rare} & {\textbf{(x100)}} \\
\midrule
Structured & \xmark & 6.20 & 8.40 & 5.54 & 26.89 \\
\rowcolor{ourtablecolor} CoT & \cmark  & 10.24 & 11.04 & 9.71 & 31.18 \\
Descriptive & \cmark  & 10.90 & 11.60 & 10.69 & 33.53 \\
\rowcolor{ourtablecolor} Direct & \cmark  & \textbf{13.46} & \textbf{15.12} & \textbf{12.96} & \textbf{39.91} \\
\bottomrule
\multicolumn{6}{c}{``PP'' stands for our post-processing pipeline (\cref{sec:method}).}
\end{tabular}%
}
\label{tab:abl_prompts_AVG}
\vspace{-2em}
\end{table}

As seen in Tab. \ref{tab:abl_visual_augmentation_lmm}, the blind model performs the worst across mAP and SR, indicating that removing visual information harms the model's ability to recognize interactions.
Red circle, reverse blur, crop with mask and crop all outperform blind, with crop being the most effective technique overall, yielding the highest mAP and $SR$.
Notably, rare interactions are particularly impacted by the choice of visual augmentation, with crop showing the best performance in this regard.
Furthermore, removing the visual context given by the image area between the human and object (\ie \textit{crop with mask} prompting) results in performance degradation. This highlights the importance of the visual context between the person and the object when predicting the interaction.

\noindent \textbf{On different text prompting strategies}.
Alongside the analysis on visual prompting, we explored different textual prompts to condition the \lmms.
We tested \textit{a)} a \textit{structured} prompt, where we ask the model to format its reply as a list of $\langle \text{\textit{human}}, \text{\textit{verb}}, \text{\textit{object}} \rangle$ triplets (note that with such prompting, the post-processing pipeline is no longer needed, as the output is already in the HOI triplet format); \textit{b)} a \textit{CoT} (Chain-of-Thought) prompt, where we append the phrase $\text{``\textit{think step-by-step}''}$ to the prompt to encourage reasoning; \textit{c)} a \textit{descriptive} prompt, where we instruct the model to describe the image while focusing on the object in the prompt; d) a \textit{direct} prompt that instructs to answer to the prompt without any additional explanation. 
For the exact prompts used, refer to \cref{sec:LLMS_det}.
The results presented in~\cref{tab:abl_prompts_AVG} show the performance on LLaVA 0.5B, with the structured prompt obtaining the worst results out of all four prompts, which also highlights the effectiveness of our post-processing pipeline.
Direct prompting achieves the best results on all metrics, with a best improvement of +3.52\% on rare mAP over the second-best prompt (descriptive).
In~\cref{sec:discuss} (\cref{tab:abl_prompts} of \supmat), we extend the evaluation to all \lmms used in this paper, with the direct prompting achieving the best results on bigger \lmms as well.

\noindent \textbf{Additional discussions.} In \cref{sec:discuss} of \supmat we show how visual instruction tuning on \hicodet and \vghoi can improve performance on \task. Furthermore, we investigate common failure modes of \lmms.%

\subsection{Qualitative results}
\label{sec:qualitatives}
In \cref{fig:main_qualitatives} and \cref{fig:supp_qualitatives,fig:supp_qualitatives_big} of \supmat we show qualitative results of LLAVA OV 0.5B~\cite{li2024llava} + \method + TT and CLIP~\cite{Radford2021}.
These results show that, even with a small 0.5B LLaVA-OV, \method produces HOI triplets that better reflect the semantics of real scenes.

\subsection{U-HOI Findings}
\label{sec:what_uhoi_reveals}
Our analysis reveals several key patterns about \lmm capabilities and limitations for unconstrained HOI detection:
\begin{itemize}
    \item \textbf{Prior HOI methods} heavily rely on verb-object co-occurrence statistics. For instance, when the occurrence matrix $\mathcal{M}$ is removed from DHD~\cite{Wu2024}, performance drops substantially (\cref{fig:plot_dhd_preliminary} in \supmat). This poses a limitation in achieving truly in-the-wild HOI detection.
    \item \textbf{Direct prompting with text-to-graph refinement outperforms structured extraction.} Qwen2-VL 7B leads with 14.68\% Full mAP under direct prompting (\cref{tab:abl_prompts} in \supmat), while structured prompts that bypass post-processing yield the worst results (6.20\% mAP, \cref{tab:abl_prompts_AVG}), highlighting the importance of our text-to-graph refinement pipeline.
    \item \textbf{Test-time compute yields substantial gains with minimal overhead.} With only 2 generations per sample, LLaVA OV 0.5B improves by +2\% mAP (\cref{fig:abl_test_time} of \supmat); scaling to $k$=10 triplets pushes performance to 20.75\% Full mAP (\cref{fig:abl_test_time}), demonstrating that sampling diversity effectively captures multiple interactions.
    \item \textbf{Visual instruction tuning transfers across datasets.} LoRA finetuning on \hicodet boosts LLaVA OV 0.5B from 13.46\% to 27.06\% mAP on \hicodet (\cref{tab:abl_finetuning_hicodet} in \supmat), and from 12.76\% to 21.04\% mAP on \vghoi (\cref{tab:abl_finetuning_vghoi} in \supmat), indicating that interaction understanding generalizes beyond the training domain.
\end{itemize}

\section{Conclusions}
\label{sec:conclusions}

We introduced \taskacro (\task), a new HOI task that removes any reliance on a predefined interaction vocabulary at training and inference. We showed the limitations of existing open-vocabulary HOI methods under this setting and propose \method, a pipeline that uses \lmms to produce free-form outputs which are then converted into HOI triplets via a post-generation procedure. We further enhanced predictions with test-time computing, which consistently improved the performance of smaller \lmms. Extensive experiments, using both standard and newly proposed metrics, demonstrate the effectiveness of \lmms for \task. Future work will explore fine-tuning, object grounding, and more advanced test-time approaches.

\section*{Acknowledgments}
\small We acknowledge EuroHPC Joint Undertaking for awarding us access to MareNostrum5 at BSC, Spain. This work was supported by the EU Horizon ELIAS (No. 101120237), ELLIOT (No. 101214398), TURING (No. 101215032), IAMI (No. 101168272), and PATTERN (No. 101159751) projects. This work was carried out in the Vision and Learning joint laboratory of FBK and UniTN.

{\small
\bibliographystyle{ieee}
\bibliography{egbib}
}

\clearpage
\setcounter{page}{1}
\maketitlesupplementary

\noindent In this document we first analyze the open-vocabulary HOI paradigm described in~\cref{sec:preliminaryDHD}, highlighting how the use of verb priors impacts performance and how omitting them leads to a major decline.
In~\cref{sec:map}, we provide additional details on the mAP metric applied to \task.
In~\cref{sec:LLMS_det}, we summarize the \lmms and the prompts used throughout this paper.
Following this, in~\cref{sec:discuss} we present an extensive study on \lmms for HOI, which could inspire future research in this area. Specifically, we aimed to explore 
which verbs are better predicted by them and investigating other related aspects. 
Furthermore, \cref{sec:additional_impl_det} includes additional implementation details,
while in~\cref{sec:supp_qualitatives} we provide a qualitative analysis comparing \method with VLMs. 
Finally, in \cref{sec:ethics}, we provide an ethics statement.

\section{Preliminary analysis on open-vocabulary HOI}
\label{sec:preliminaryDHD}

\begin{figure}[!ht]
    \centering
    \includegraphics[width=1.0\linewidth]{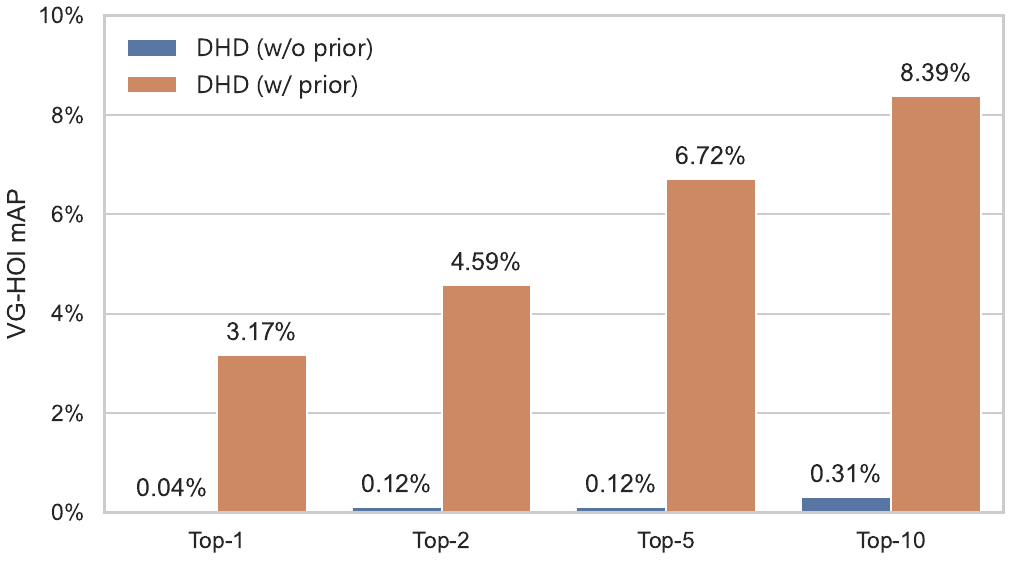}
    \caption{Evaluation of DHD~\cite{Wu2024} on the \vghoi dataset with and without the occurrence matrix $\mathcal{M}$.}
    \label{fig:plot_dhd_preliminary}
\end{figure}

In Fig.~\ref{fig:plot_dhd_preliminary}, we present the performance of a state-of-the-art open-set HOI method, DHD~\cite{Wu2024}, on the \vghoi dataset. 
Specifically, we evaluate %
DHD~\cite{Wu2024} both with sparse co-occurrence matrix $\mathcal{M} \in \{0,1\}^{|\mathcal{O}| \times |\mathcal{V}|}$, which defines the set of permissible verb-object interactions in a given dataset (\ie, as the method was originally designed) and without it. This experiment highlights that such models are not truly open-vocabulary, as their performance drops dramatically when prior knowledge of verb occurrence, derived solely from the frequency of interactions in the dataset, is removed.
We argue that open-vocabulary models should not rely on such prior knowledge but should instead function independently, without constraints or external aids.

\section{Description of mAP}
\label{sec:map}
Herein, we describe mAP to help readers better understand its rationale and facilitate the understanding of the changes needed for \task.

In earlier HOI detection works, \eg, \cite{zhang2023exploring,Ning2023,Lei2023,Liao2022,Yang24,Wu2024}, the verb categories are restricted to a fixed vocabulary $\mathcal{V}$.
Let $\mathcal{P} = {(b_{h_i}, b_{o_i}, v_i, s_i)}_{i=1}^{N}$ denote the set of $N$ predicted HOI triplets for an image, where $b_{h_i} \in \mathbb{R}^4$ and $b_{o_i} \in \mathbb{R}^4$ denote the human and object predicted bounding boxes, $v_i \in \mathcal{V}$ is the interaction verb, and $s_i \in [0, 1]$ is its associated confidence score.
Ground-truth annotations $\mathcal{G} = {(b_{h_j}, b_{o_j}, v_j)}_{j=1}^{M}$ are also provided with $v_j \in \mathcal{V}$.

Under this setting, predictions are first matched to ground-truth instances based on spatial alignment (typically an IoU criterion, \eg IoU~$\geq 0.5$) and exact category correspondence.
For each verb category, the Average Precision (AP) is computed as the area under the precision-recall curve, and the overall performance is summarized by the mean Average Precision (mAP), obtained by averaging the APs over all categories in $\mathcal{V}$.
Benchmarks like HICODET~\cite{Chao2018}, also compute aggregated results depending on the rarity of the interactions (\ie, Rare and Non-rare).
  
Since we aim to predict interaction verbs from an unbounded vocabulary $\Omega$ (with $\mathcal{V} \subset \Omega, \quad |\Omega| \gg |\mathcal{V}|$), the conventional mAP, computed over a fixed verb set $\mathcal{V}$, is no longer directly applicable.
To evaluate \task detectors under our setting, we adapt the mAP to match predicted and ground truth verbs based on the embedding semantic similarity.

Let $\mathcal{P}' = \{(b_{h_i}, b_{o_i}, v'_i)\}_{i=1}^{N'}$ denote the set of predicted HOI triplets for an image, where $b_{h_i} \in \mathbb{R}^4$ and $b_{o_i} \in \mathbb{R}^4$ denote the human and object bounding boxes, respectively, and $v'_i \in \Omega$ is the interaction verb.
To match these unconstrained predictions with the verb evaluation set~$\mathcal{V}$, we follow the approach of~\cite{li2023factual, diko2024semantically} and employ a BERT-based~\cite{bert} similarity function $\mathrm{sim}(\cdot, \cdot)$ that measures the semantic similarity between a predicted verb $v'_i$ and each verb $v \in \mathcal{V}$.
Therefore, for a given similarity threshold $\tau$, we associate $v'_i$ with any $v \in \mathcal{V}$ that satisfies $\mathrm{sim}(v'_i, v) \ge \tau$,
\begin{equation}
    \mathrm{sim}(v'_i, v) \ge \tau,
\end{equation}
thereby forming a new prediction set $\mathcal{P}^{\tau}$ where the verbs strictly belong to $\mathcal{V}$.
Furthermore, inspired by object detection evaluation~\cite{Lin2014}, which accounts for different degrees of localization uncertainty, we compute the mAP across multiple thresholds $\mathcal{T}$ (see \cref{sec:impdet} for the selected thresholds).
The final mAP score is then computed as the average mAP for each threshold $\mathcal{T}$.
\begin{equation}
    \text{mAP Avg.} = \frac{1}{|\mathcal{T}|} \sum_{\tau \in \mathcal{T}} \text{mAP}(\mathcal{P}^{\tau}).
\end{equation}

Given that current HOI datasets are not exhaustively labeled, it is not possible to determine whether a missing verb in the ground truth is due to an annotation omission or a genuine error by the model. Therefore, coherently with open-world object detection and tracking~\cite{Liu22,Zhao24} our benchmark does not explicitly penalize false positives.

\section{Details of \lmms and prompts used}
\label{sec:LLMS_det}
We summarize the \lmms used in this study in~\cref{tab:supp_models_info}.
In~\cref{tab:supp_prompts} we provide the exact prompts used in the experiments, and Tab.~\ref{tab:abl_prompts} extends the results shown in the main paper (\cref{sec:ablations}) for all tested models.

\begin{table}[]
\caption{Summary of the vision and language towers of the \lmms we tested in this work.}
\centering
\resizebox{\linewidth}{!}{
\begin{tabular}{l|ll|ll}
    \toprule
    \multicolumn{1}{c|}{\textbf{Method}} & \multicolumn{2}{c|}{\textbf{Vision Tower}} & \multicolumn{2}{c}{\textbf{Text Tower}} \\
    \midrule
    {} & \textbf{Model} & \textbf{Num. params.} & \textbf{Model} & \textbf{Num. params.} \\
    \rowcolor{ourtablecolor} {Idefics2} \cite{laurenccon2025matters} & SigLIP & 400M & Mistral & 7B \\
    {InstructBLIP} \cite{huang2023visual} & ViT-g (BLIP-2) & 1.1B & Vicuna & 7B/13B \\
    \rowcolor{ourtablecolor} {InternVL2} \cite{chen2024internvl} & InternViT & 300M/6B & Qwen2 500MB or InternLM2 8B & 500M/8B  \\
    {LLaVA-OV} \cite{li2024llava} & SOViT (SigLIP) & 400M & Qwen2 & 500M/7B \\
    \rowcolor{ourtablecolor} {Phi3V} \cite{abdin2024phi} & ViT-L (CLIP) & 400M & 3.8B & MIT \\
    {Qwen2VL} \cite{wang2024qwen2} & ViT & 600M & Qwen2 & 1.5B/7B \\
    \bottomrule
\end{tabular}}
\label{tab:supp_models_info}
\end{table}

\begin{table}[]
\caption{The prompts used in the ablation study in Sec. \ref{sec:experiments}.}
\centering
\resizebox{1.0\linewidth}{!}{
\begin{tabular}{l|l}
\toprule
\multicolumn{2}{l}{Prompts} \\
\midrule
\rowcolor{ourtablecolor} {} & What are the interactions between the person \\
\rowcolor{ourtablecolor} \multirow{-2}{*}{Direct} & and the \texttt{obj}? \\
\multirow{2}{*}{CoT} & What are the interactions between the person \\
{} & and the \texttt{obj}? Think step by step. \\
\rowcolor{ourtablecolor} {} & Describe all the interactions occurring between the person \\
\rowcolor{ourtablecolor} {} & and the \texttt{obj}. If no interactions are being performed \\
\rowcolor{ourtablecolor} \multirow{-3}{*}{Descriptive} & reply with ``no interaction''. \\
\multirow{4}{*}{Structured} & What are the interactions between the person and the \texttt{obj}? \\
{} & Answer the question with triplets of the form: (person, verb, \texttt{obj}).  \\
{} & Examples: ``(person, sit on, bike). (person, ride, bike)''. If there are no \\ 
{} & interactions, answer with (person, none, \texttt{obj}). Do not write any other text. \\
\bottomrule
\end{tabular}%
}
\label{tab:supp_prompts}
\end{table}

\begin{table}[!t]
\caption{Comparison of \method under different prompts on the HICO-DET~\cite{Chao2018} dataset and the \publicsetting setting. Best results are in \textbf{bold}.}
\centering
\resizebox{0.85\columnwidth}{!}{%
\begin{tabular}{@{}cl|ccc|c@{}}
\toprule
\multicolumn{2}{c|}{\multirow{2}{*}{\textbf{Method}}}  & \multicolumn{3}{c|}{\textbf{mAP Avg. (\%)} $\uparrow$} &  \multicolumn{1}{c}{\textbf{SR} $\mathbf{\uparrow}$}  \\
& & \multicolumn{1}{c}{Full} & \multicolumn{1}{c}{Rare} & \multicolumn{1}{c|}{Non-rare} &  \multicolumn{1}{c}{\textbf{(x100)}} \\
\midrule
\multicolumn{1}{l}{\multirow{9}{*}{\rotatebox[origin=c]{90}{CoT}}} & Idefics2 8B~\cite{laurenccon2025matters} & 14.04 & 14.96 & 13.77 & 42.87 \\
\multicolumn{1}{c}{}                       & InternVL 2B~\cite{chen2024internvl}  & 8.46 & 9.72 & 8.08 & 28.51 \\
\multicolumn{1}{c}{}                       & InternVL 8B~\cite{chen2024internvl}  & 9.40 & 9.82 & 9.27 & 30.21 \\
\multicolumn{1}{c}{}                       & InstructBLIP 7B~\cite{huang2023visual}  & 11.46 & 14.52 & 10.54 & 41.68 \\
\multicolumn{1}{c}{}                       & InstructBLIP 13B~\cite{huang2023visual}  & 11.12 & 13.32 & 10.47 & 42.70 \\
\multicolumn{1}{c}{}                       & LLaVA OV 0.5B~\cite{li2024llava}  & 13.34 & 14.36 & 13.04 & 31.18 \\
\multicolumn{1}{c}{}                       & LLaVA OV 7B~\cite{li2024llava}  & 12.36 & 14.19 & 11.82 & 15.00 \\
\multicolumn{1}{c}{}                       & Qwen2-VL 2B~\cite{wang2024qwen2}  & 4.12 & 4.14 & 4.11 & 17.02 \\
\multicolumn{1}{c}{}                       & Qwen2-VL 7B~\cite{wang2024qwen2}  & 7.72 & 7.45 & 7.81 & 22.25 \\
\midrule
\multicolumn{1}{l}{\multirow{9}{*}{\rotatebox[origin=c]{90}{Descriptive}}} & Idefics2 8B~\cite{laurenccon2025matters} & 7.28 & 9.22 & 6.70 & 14.95 \\
\multicolumn{1}{c}{}                       & InternVL 2B~\cite{chen2024internvl}  & 10.86 & 8.86 & 11.46 & 35.21 \\
\multicolumn{1}{c}{}                       & InternVL 8B~\cite{chen2024internvl}  & 12.62 & 13.31 & 12.42 & 39.85 \\
\multicolumn{1}{c}{}                       & InstructBLIP 7B~\cite{huang2023visual}  & 12.37 & 13.22 & 12.12 & 39.84 \\
\multicolumn{1}{c}{}                       & InstructBLIP 13B~\cite{huang2023visual}  & 11.81 & 13.77 & 11.22 & 43.57 \\
\multicolumn{1}{c}{}                       & LLaVA OV 0.5B~\cite{li2024llava}  & 10.89 & 11.60 & 10.68 & 33.53 \\
\multicolumn{1}{c}{}                       & LLaVA OV 7B~\cite{li2024llava}  & 9.69 & 11.73 & 9.08 & 38.13 \\
\multicolumn{1}{c}{}                       & Qwen2-VL 2B~\cite{wang2024qwen2}  & 3.43 & 2.66 & 3.66 & 7.58 \\
\multicolumn{1}{c}{}                       & Qwen2-VL 7B~\cite{wang2024qwen2}  & 13.23 & 15.03 & 12.69 & 37.22 \\
\midrule
\multicolumn{1}{l}{\multirow{9}{*}{\rotatebox[origin=c]{90}{Direct question}}} & Idefics2 8B~\cite{laurenccon2025matters} & 10.28 & 10.15 & 10.32 & 34.27 \\
\multicolumn{1}{c}{}                       & InternVL 2B~\cite{chen2024internvl}  & 10.21 & 8.87 & 10.61 & 36.59 \\
\multicolumn{1}{c}{}                       & InternVL 8B~\cite{chen2024internvl}  & 11.25 & 10.68 & 11.42 & 37.82 \\
\multicolumn{1}{c}{}                       & InstructBLIP 7B~\cite{huang2023visual}  & 11.53 & 13.02 & 11.09 & 40.52 \\
\multicolumn{1}{c}{}                       & InstructBLIP 13B~\cite{huang2023visual}  & 11.09 & 12.93 & 10.54 & 43.57 \\
\multicolumn{1}{c}{}                       & LLaVA OV 0.5B~\cite{li2024llava}  & 13.46 & 15.12 & 12.96 & 39.91 \\
\multicolumn{1}{c}{}                       & LLaVA OV 7B~\cite{li2024llava}  & 13.01 & 14.12 & 12.68 & 39.89 \\
\multicolumn{1}{c}{}                       & Qwen2-VL 2B~\cite{wang2024qwen2}  & 12.89 & 13.45 & 12.72 & 41.21 \\
\multicolumn{1}{c}{}                       & Qwen2-VL 7B~\cite{wang2024qwen2}  & 14.68 & 16.92 & 14.02 & 45.74 \\
\bottomrule  
\end{tabular}%
}
\label{tab:abl_prompts}
\end{table}

\section{Additional Discussions}
\label{sec:discuss}

\noindent \textbf{Different text prompting strategies}.
In~\cref{tab:abl_prompts} of \supmat, we extend the evaluation of different text prompts (\cref{sec:ablations}) to all \lmms used in this paper.
In line with the results of~\cref{tab:abl_prompts_AVG} of the main paper, 
direct prompting achieves the best overall performance
with Qwen2-VL 7B being the best model, achieving a Full mAP performance of 14.68\%, followed by descriptive (13.23\%) and CoT (7.72\%). 
Other models, such as Idefics2 8B (mAP 10.28\%), also show competitive results under this strategy. 
Despite encouraging step-by-step reasoning, CoT prompts do not consistently improve performance.
Several models, like Qwen2-VL 2B (mAP 4.12\%), struggle significantly under CoT. Besides, Idefics2 8B drops from 10.28\% (Direct) to 7.28\% (Descriptive), and Qwen2-VL 2B performs the worst overall (3.43\% mAP). However, a few models, like Qwen2-VL 7B (13.23\% mAP), perform relatively well with descriptive prompts.

\noindent \textbf{The impact of $\pmb{k}$.}
We analyzed the impact of $k$ (\ie, the number of interaction triplets considered at test time) on the performance of CLIP~\cite{Radford2021} and LLaVA OV 0.5B~\cite{li2024llava} + \method, evaluating them with $k= \{1, 2, 5, 10\}$.
The results in~\cref{tab:abl_k} include all proposed metrics and are based on the HICO-DET~\cite{Chao2018} on the \publicsetting setting. Increasing $k$ generally leads to improved performance on mAP and $SR$, indicating that incorporating multiple candidate triplets enhances robustness and recall.
\begin{table}[]
\caption{The impact of $k$, \ie, the number of interaction triplets at test time on the HICO-DET~\cite{Chao2018} dataset and the \publicsetting setting.}
\centering
\resizebox{0.65\columnwidth}{!}{%
\begin{tabular}{cc|c|ccc|c}
\toprule
\multicolumn{2}{c|}{\multirow{2}{*}{\textbf{Method}}} & \multirow{2}{*}{\textbf{$k$}} & \multicolumn{3}{c|}{\textbf{mAP Avg. (\%)} $\uparrow$} & \textbf{SR} $\mathbf{\uparrow}$   \\
& {} & {} & Full & Rare & Non-rare & {\textbf{(x100)}}  \\
\midrule
\multirow{4}{*}{\rotatebox[origin=c]{90}{CLIP~\cite{Radford2021}}} & \multirow{4}{*}{\vghoi} & 1 & 2.09 & 5.44 & 1.10 & 21.79   \\
{} & {}                      & 2  & 3.11 & 7.77 & 1.71 & 26.60  \\
{} & {}                      & 5  & 4.51 & 10.70 & 2.66 & 32.80  \\
{} & {}                      & 10  & 5.49 & 11.47 & 3.44 & 37.55  \\
\rowcolor{ourtablecolor} {} & {} & 1 & 1.32 & 3.16 & 0.76 & 21.25  \\
\rowcolor{ourtablecolor} {} & {}                      & 2  & 2.41 & 6.47 & 1.19 & 25.49  \\
\rowcolor{ourtablecolor} {} & {}                      & 5  & 3.60 & 9.31 & 1.90 & 30.37  \\
\rowcolor{ourtablecolor} {\multirow{-4}{*}{\rotatebox[origin=c]{90}{CLIP~\cite{Radford2021}}}} & {\multirow{-4}{*}{WordNet~\cite{miller1995wordnet}}}                      & 10  & 4.51 & 11.24 & 2.50 & 33.86  \\
\multicolumn{2}{l|}{\multirow{4}{*}{LLaVA OV 0.5B~\cite{li2024llava}}} & 1 & 14.81 & 16.17 & 14.40 & 45.76 \\
{} & {}                      & 2  & 18.42 & \textbf{33.11} & 17.46 & 53.54 \\
{} & {}                      & 5  & 20.31 & 23.92 & 19.24 & 60.96  \\
{} & {}                      & 10  & \textbf{20.75} & 25.25 & \textbf{19.40} & \textbf{63.54} \\
\bottomrule  
\end{tabular}%
}
\label{tab:abl_k}
\end{table}

\noindent \textbf{Effects of visual instruction tuning for \task}.
In \cref{tab:abl_finetuning_hicodet,tab:abl_finetuning_vghoi}, we present the gains of visual instruction tuning for HOI.
More specifically, we added LoRA modules to the vision tower and projector of LLaVA OV 0.5B~\cite{li2024llava} and finetuned them on the training set of \hicodet.
The results show that finetuning can greatly improve the results on both \hicodet and \vghoi datasets.

\begin{table}[!t]
\caption{Performance of \method on LLaVA OV 0.5B on \hicodet using the pre-trained weights or after a LoRA finetuning stage with the \hicodet \publicsetting setting training set. Best results are in \textbf{bold}.}
\centering
\resizebox{0.8\columnwidth}{!}{%
\begin{tabular}{@{}cl|ccc|c@{}}
\toprule
\multicolumn{1}{c}{\multirow{2}{*}{\textbf{Setting}}} & \multicolumn{1}{c|}{\multirow{2}{*}{\textbf{Method}}} & \multicolumn{3}{c|}{\textbf{mAP Avg. (\%)} $\uparrow$} & \textbf{SR} $\mathbf{\uparrow}$  \\
& & \multicolumn{1}{c}{Full} & \multicolumn{1}{c}{Rare} & \multicolumn{1}{c|}{Non-rare} & {\textbf{(x100)}} \\
\midrule
\rowcolor{ourtablecolor} \multicolumn{1}{l}{} & LLaVA OV 0.5B~\cite{li2024llava}  & 13.46 & 15.12 & 12.96 & 39.91 \\
\rowcolor{ourtablecolor} \multicolumn{1}{l}{\multirow{-2}{*}{Annotated-box}} & + LoRA~\cite{hu2022lora}  & \textbf{27.06} & \textbf{17.02} & \textbf{30.06} & \textbf{58.64} \\
\multicolumn{1}{l}{\multirow{2}{*}{Computed-box}} & LLaVA OV 0.5B~\cite{li2024llava}  & 6.43 & 15.12 & 12.96 & 36.25 \\
\multicolumn{1}{l}{} & + LoRA~\cite{hu2022lora}  & \textbf{16.72} & \textbf{10.89} & \textbf{18.46} & \textbf{43.65} \\ 
\bottomrule  
\end{tabular}%
}
\label{tab:abl_finetuning_hicodet}
\end{table}

\begin{table}[!t]
\caption{Performance of \method on LLaVA OV 0.5B on \vghoi using the pre-trained weights or after a LoRA finetuning stage with the \hicodet \publicsetting setting training set. Best results are in \textbf{bold}.}
\centering
\resizebox{0.8\columnwidth}{!}{%
\begin{tabular}{@{}cl|c|c@{}}
\toprule
\textbf{Setting} & \multicolumn{1}{c|}{{\textbf{Method}}}  & \textbf{mAP Avg. (\%)} $\uparrow$ & \textbf{SR (x100)} $\mathbf{\uparrow}$  \\
\midrule
\rowcolor{ourtablecolor} \multicolumn{1}{l}{} & LLaVA OV 0.5B~\cite{li2024llava} & 12.76 & 42.70  \\
\rowcolor{ourtablecolor} \multicolumn{1}{l}{\multirow{-2}{*}{Annotated-box}} & + LoRA~\cite{hu2022lora}  & \textbf{21.04} & \textbf{66.49} \\
\multicolumn{1}{l}{\multirow{2}{*}{Computed-box}} & LLaVA OV 0.5B~\cite{li2024llava}  & 1.99 & 8.10 \\
\multicolumn{1}{l}{} & + LoRA~\cite{hu2022lora}  & \textbf{2.76} & \textbf{12.55} \\ 
\bottomrule  
\end{tabular}%
}
\label{tab:abl_finetuning_vghoi}
\end{table}

\begin{figure}[!t]
    \centering
    \includegraphics[width=1.0\linewidth]{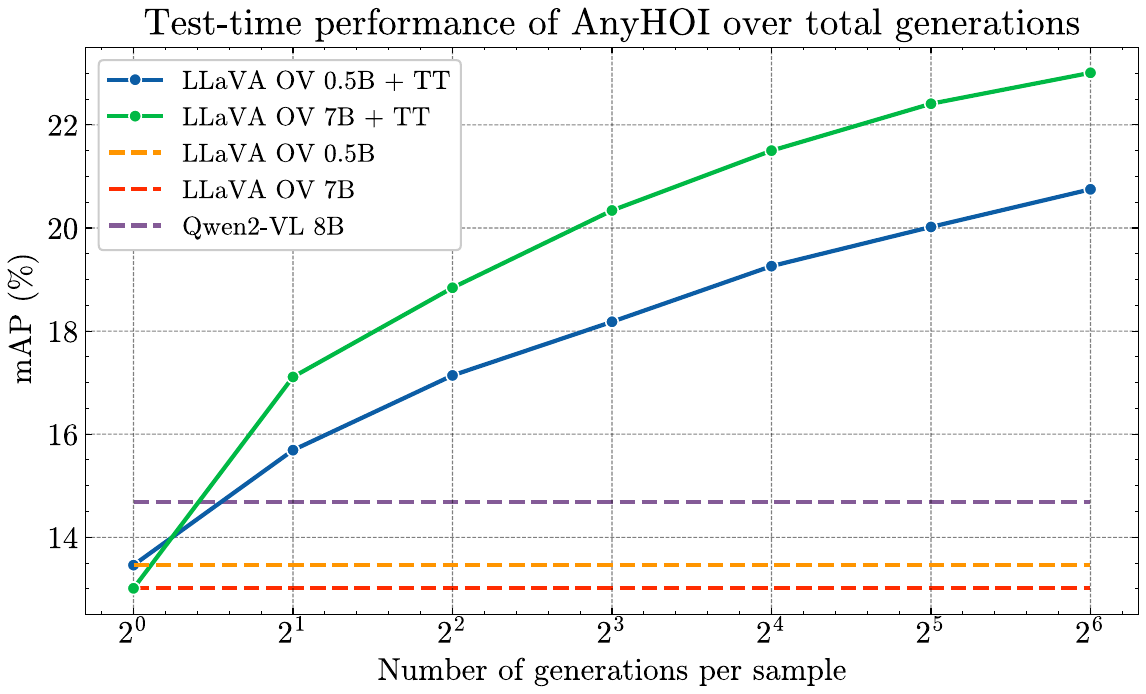}
    \caption{Results of LLaVA + \method + TT varying the number of generations on the \hicodet dataset under the \publicsetting setting. Other methods using \method without TT are plotted for reference.}
    \label{fig:abl_test_time}
    \vspace{-1.2em}
\end{figure}

\noindent \textbf{\method with increasing number of generations.}
In Fig.~\ref{fig:abl_test_time}, we show how performance improves as we increase the number of generated answers on \method + TT.
Note that with just 2 generations per sample, LLaVA 0.5B results already improve by +2\% mAP, and +6\% with 4 generations. This demonstrates that test-time sampling with a temperature of 0.2 enables exploration of different generation paths in the \lmms, potentially capturing multiple distinct interactions.

\noindent \textbf{Performance of \lmms on verbs grouped by topic.}  
We analyzed the performance of \lmms on verbs categorized by the topics listed below. We utilized GPT4-o to assign verbs to the most suitable topics. The analysis have been made on 
\hicodet dataset for the \publicsetting setting.

\begin{itemize}  
    \item \textbf{Opposite verbs}: to assess the model's capability to distinguish opposing actions, e.g., pulling vs. pushing, assembling vs. disassembling, catching vs. throwing, boarding vs. exiting, sitting vs. standing.
    \item \textbf{Social verbs}: to evaluate the model's ability to categorize verbs related to social interactions, e.g., kissing or hugging.  
    \item \textbf{Object manipulation}: to determine whether the model can identify verbs representing physical manipulation of objects, such as opening or inspecting.  
    \item \textbf{Physical activities}: to examine the model's recognition of verbs associated with physical activities, such as running, jumping, and walking.  
    \item \textbf{Verbs used with prepositions}: to investigate the model's performance on verbs that are typically used with prepositions to describe specific spatial relationships or actions involving locations, such as ``lying on'', ``sitting at'', ``standing on'', and ``standing under''.

\end{itemize}  

\begin{figure*}
    \centering
    \includegraphics[width=0.7\linewidth]{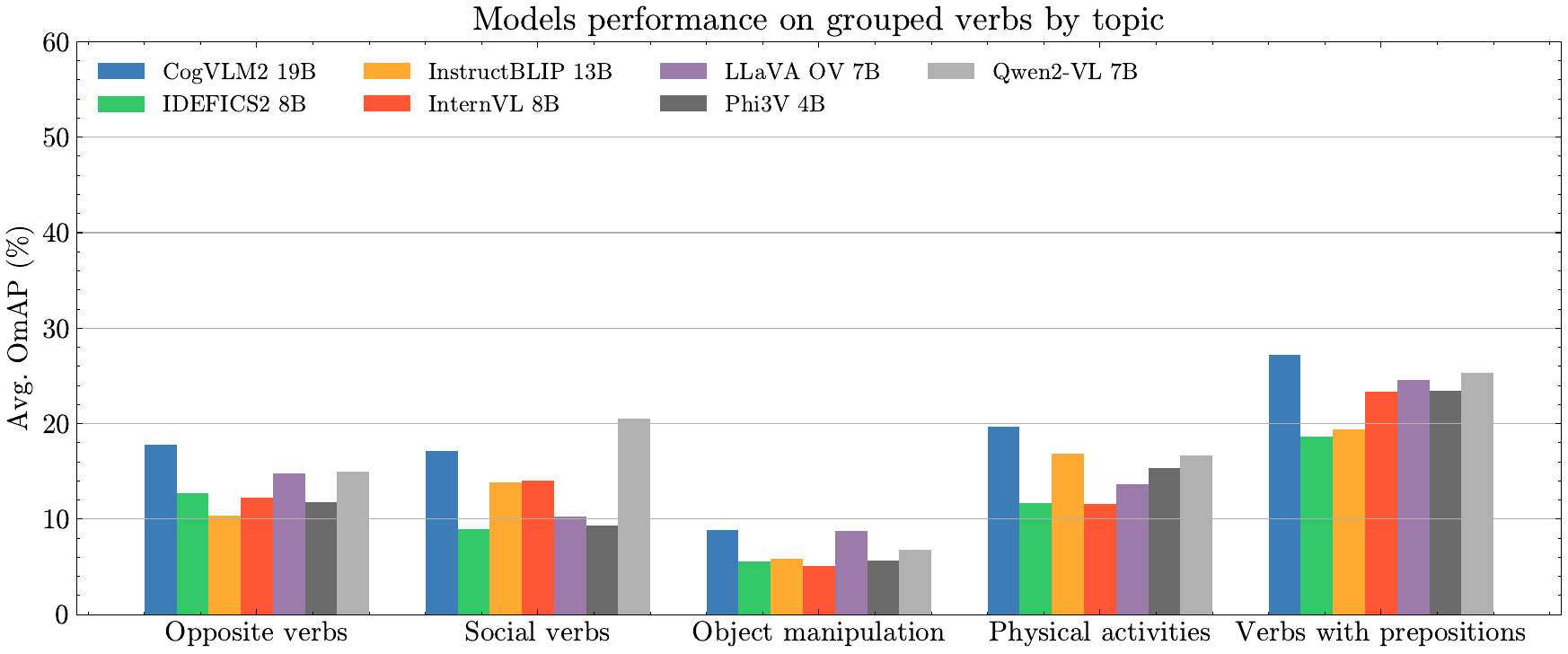}
    \caption{mAP of \lmms on grouped verbs by topic.}
    \label{fig:plot_models_by_topic}
\end{figure*}

\begin{figure*}
    \centering
    \includegraphics[width=0.7\linewidth]{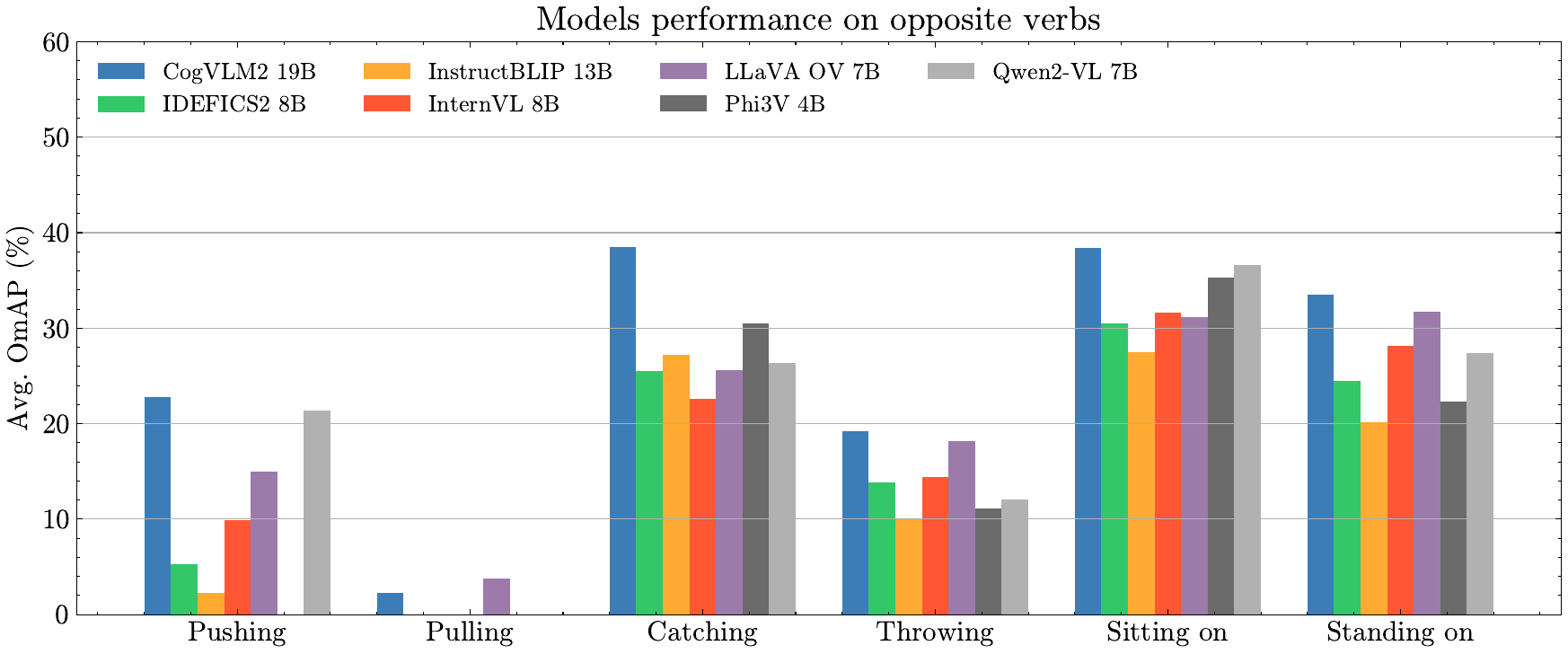}
    \caption{mAP of \lmms on opposite verbs.}
    \label{fig:plot_models_opposite_verbs}
\end{figure*}

\begin{figure*}
    \centering
    \includegraphics[width=0.7\linewidth]{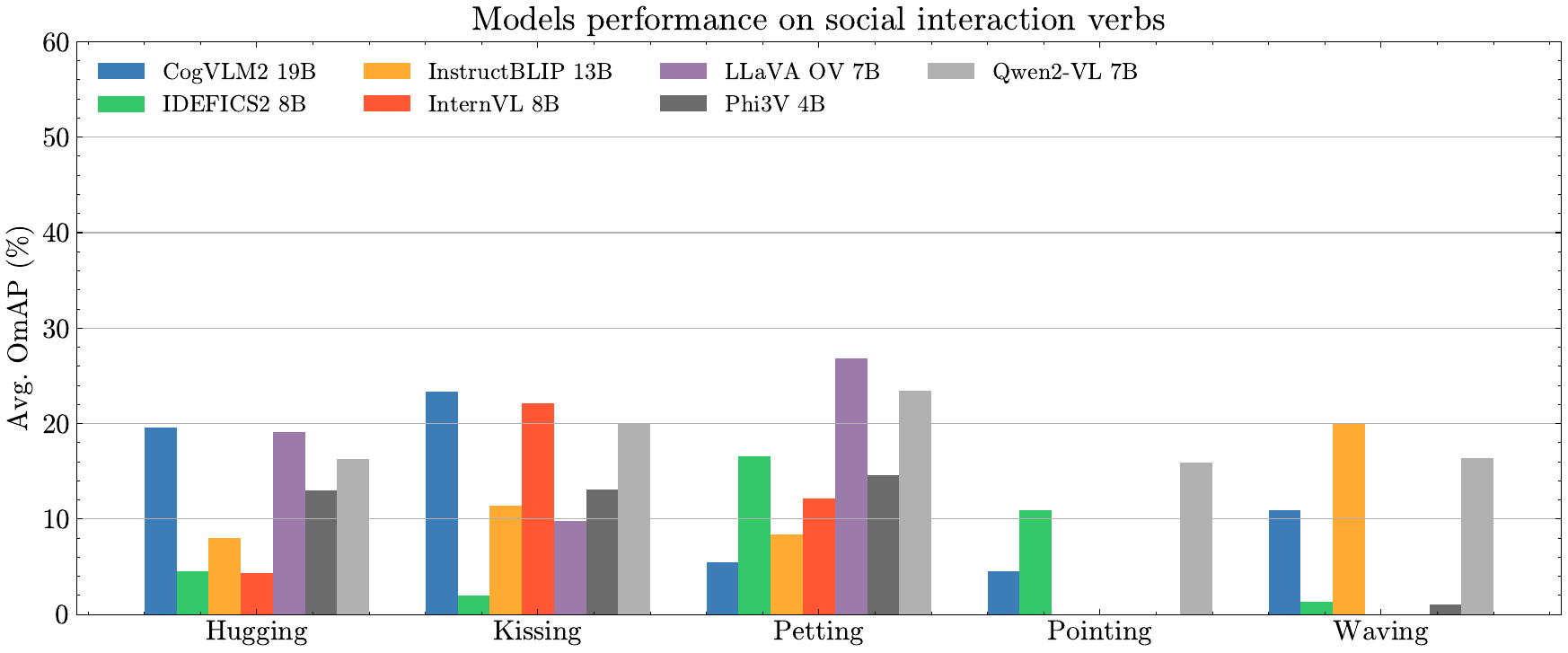}
    \caption{mAP of \lmms of social interaction verbs.}
    \label{fig:plot_models_social_interactions}
\end{figure*}

\begin{figure*}
    \centering
    \includegraphics[width=\linewidth]{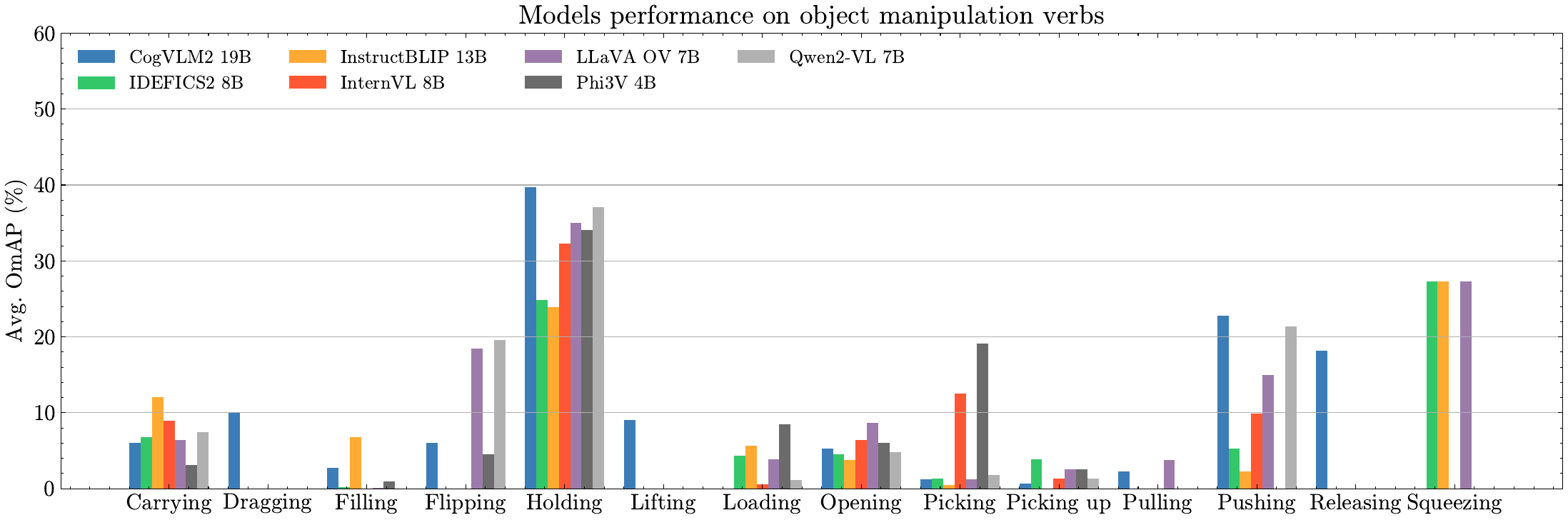}
    \caption{mAP of \lmms on object manipulation verbs.}
    \label{fig:plot_models_object_manipulation}
\end{figure*}

\begin{figure*}
    \centering
    \includegraphics[width=0.8\linewidth]{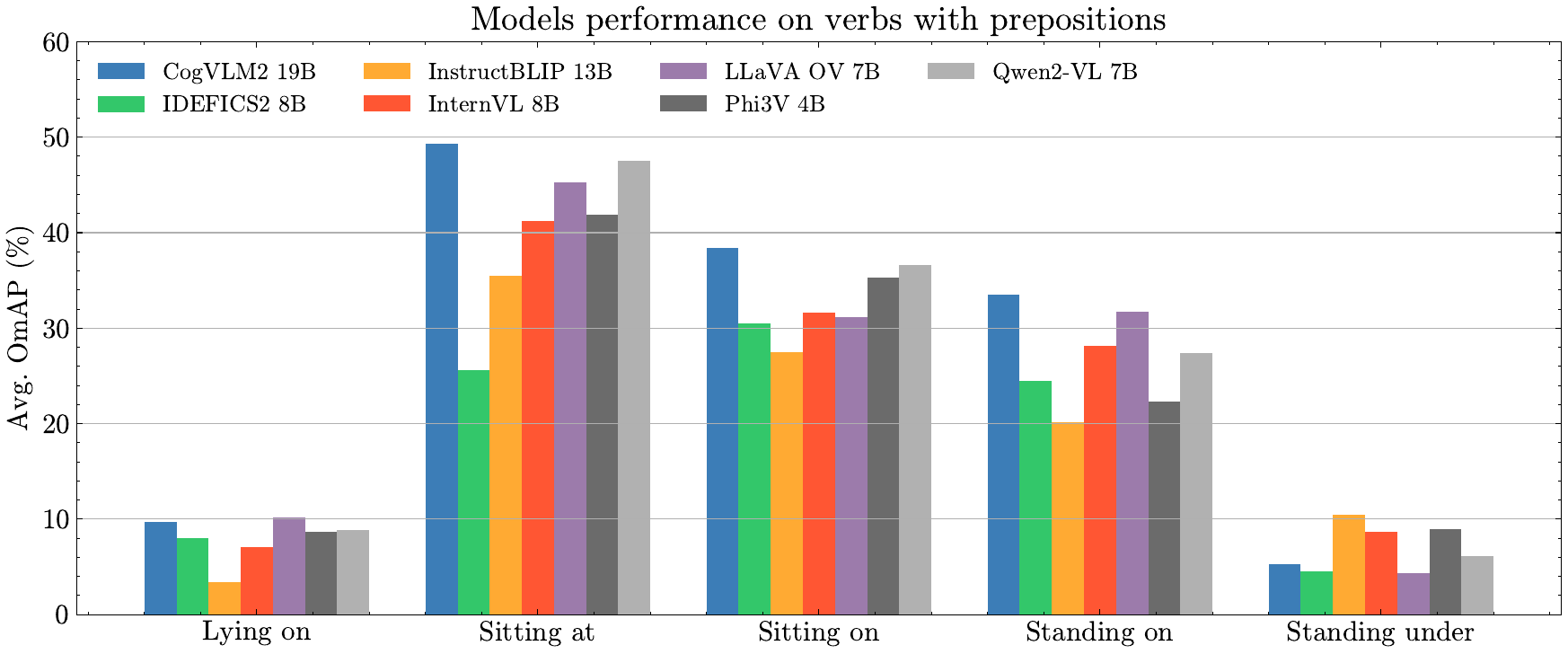}
    \caption{mAP of \lmms performance on verbs with propositions}
    \label{fig:plot_models_positional_verbs}
\end{figure*}

The corresponding results in terms of mAP are provided in Figs. \ref{fig:plot_models_by_topic}, \ref{fig:plot_models_opposite_verbs},  \ref{fig:plot_models_social_interactions}, \ref{fig:plot_models_object_manipulation}, and \ref{fig:plot_models_positional_verbs}. Fig. \ref{fig:plot_models_by_topic} shows the overall performance of \lmms across topics, Fig. \ref{fig:plot_models_opposite_verbs} shows the performance for opposite verbs, Fig. \ref{fig:plot_models_social_interactions} shows the performance for social verbs, Fig. \ref{fig:plot_models_object_manipulation} shows the performance for object manipulation, and Fig. \ref{fig:plot_models_positional_verbs} demonstrates the performance for verbs used with prepositions.

In general, when the verbs are grouped in terms of the aforementioned topics, the performances of different \lmms exhibit similar behavior. For instance, physical activities appear to be the most challenging to detect for all, followed by object manipulations. For social verbs, Idefics2 8B, InternVL, and Qwen2-VL 7B outperform the others. For opposite verbs, CogVLM2 19B is the best, followed by Qwen2-VL 7B. Among opposite verbs, the most challenging one is the ``pulling'' action, while ``pushing'' is the verb for which the agreement among different \lmms is the worst, meaning that some models perform particularly poorly.
For object manipulations, there are several verbs where all MLLM types show poorer performance, such as setting, spinning, and wielding. Moreover, verbs with prepositions typically refer to positional verbs that describe the position of a person, such as ``sitting'', ``lying'', or ``standing''. The \lmms perform well in distinguishing between clear and simple positional states, like ``sitting'' versus ``standing''. However, they face challenges when dealing with propositional verbs, which involve a relationship with objects or locations, or verbs that require more subtle distinctions. For example, verbs like ``sitting'' and ``lying'' describe similar body postures but have different contexts, which makes it harder for the models to accurately interpret them. \\

\noindent \textbf{Inference times.}~\cref{tab:abl_inference_times} shows inference time of \lmms. Like traditional HOI methods (\eg MP-HOI inference time is $<1~\text{sec.}$) \method shows competitive inference times that are acceptable for in-the-wild applications such as human-robot collaboration, where reaction times of ~1 second are considered to be acceptable~\cite{shiwa2008quickly}.
\begin{table}[!ht]
\caption{Inference times per interaction of \lmms and \method.}
\centering
\resizebox{0.6\columnwidth}{!}{%
\begin{tabular}{@{}lcc@{}}
\toprule
Model                    & Inference time (s) & Full mAP \\ \midrule
CogVLM2                  & 1.03               & 1.34     \\
+ \method                & 1.13               & 8.10     \\
\rowcolor{ourtablecolor} InternVL2-8B             & 0.77               & 1.91     \\
\rowcolor{ourtablecolor} + \method                & 0.87               & 6.47     \\
Qwen2-VL                 & 0.65               & 4.03     \\
+ \method                & 0.77               & 9.35     \\
\rowcolor{ourtablecolor} LLaVA 7B                 & 0.66               & 2.72     \\
\rowcolor{ourtablecolor} + \method                & 0.86               & 8.73     \\
LLaVA 0.5B               & 0.69               & 1.32   \\
+ \method + TT           & 1.19               & 10.30    \\ \bottomrule
\end{tabular}%
}
\label{tab:abl_inference_times}
\end{table}

\noindent \textbf{mAP with different threshold values.}
As mentioned in~\cref{sec:impdet}, we evaluated mAP with different thresholds $\mathcal{T} = \{0.6, 0.7, 0.8, 0.9, 0.95\}$.
In Tables \ref{tab:supp_hicodet_classification}, \ref{tab:supp_hicodet_detection}, \ref{tab:supp_vghoi_classification}, and \ref{tab:supp_vghoi_detection}, we present the corresponding results for \hicodet and \vghoi datasets under the \publicsetting and \privatesetting settings.

In all tables, higher thresholds that require a more precise match (e.g., @0.9, @0.95) consistently show lower mAP scores. LLaVA + \method + TT achieves competitive results compared to other approaches (e.g., CLIP, CogVLM2, InternVL, InternVL + \method) across different thresholds and categories, with notable improvements in rare HOI detection.
The model's ability to enhance detection performance in more challenging (rare) categories and at higher precision thresholds highlights its effectiveness in improving HOI detection tasks.

\section{Additional implementation details}
\label{sec:additional_impl_det}
The prompt we used to pre-process the verbs of WordNet (\cref{sec:impdet}) is the following: ``Can a person \texttt{verb} an object? Reply with either yes or no'', where \texttt{verb} is from WordNet.

\noindent \textbf{Adaptation of baselines.} Recall that due to the unconstrained nature of \task, some HOI methods cannot be applied and tested (\cref{sec:experiments}).
However, we successfully adapted the CLIP classification head of one-stage and two-stage open-vocabulary models~\cite{Yang24,Liao2022,Ning2023,li2023neural,kim2025locality} to use only verbs (i.e., with the prompt \texttt{A photo of a person verb something}, where \texttt{verb} is drawn from the verb vocabulary). This adaptation allowed us to evaluate these models on \task.

\section{Additional qualitative results}
\label{sec:supp_qualitatives}
\begin{figure*}
    \centering
    \includegraphics[width=0.9\linewidth]{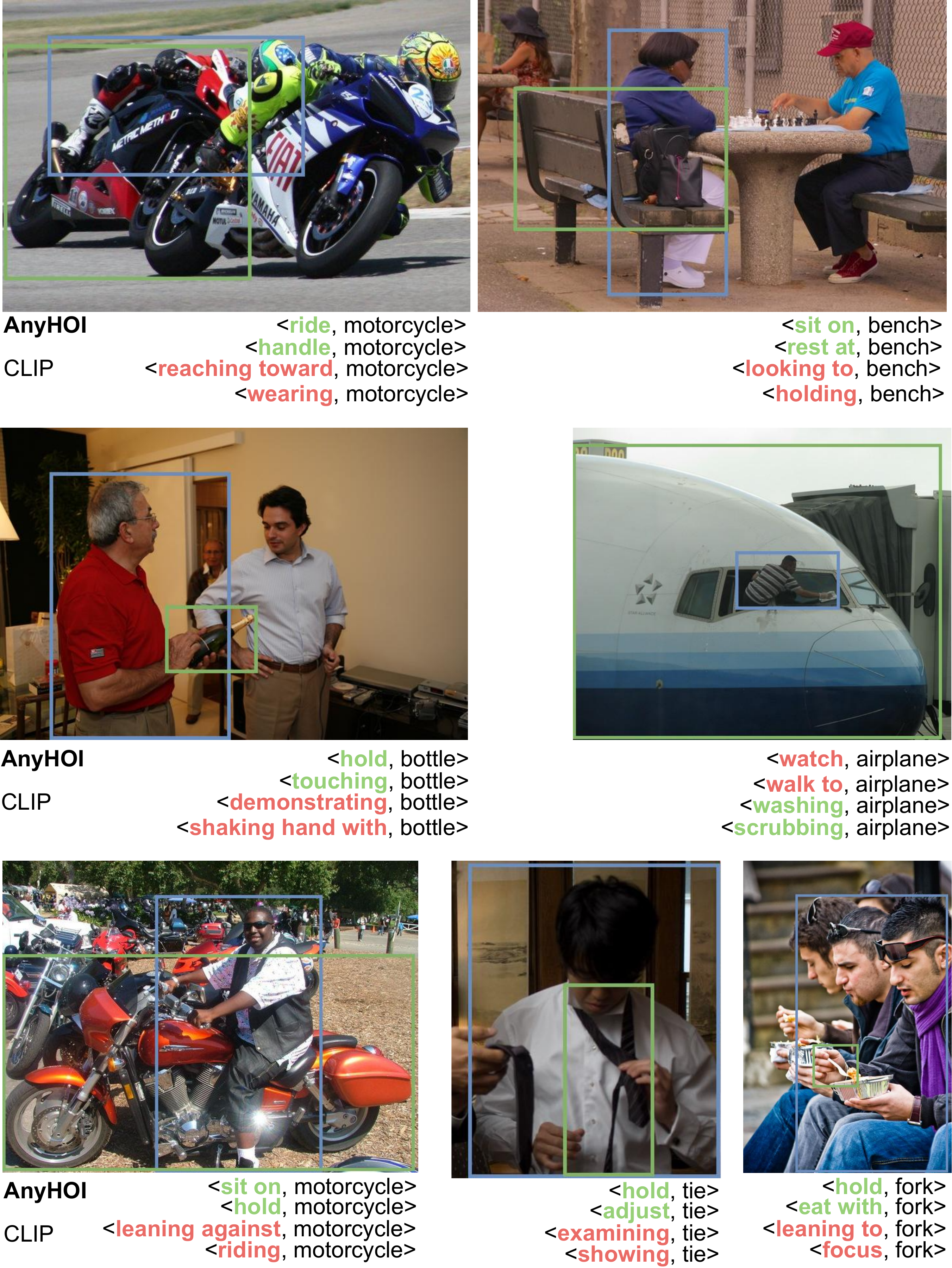}
    \caption{Qualitative results of LLaVA OV 0.5B~\cite{li2024llava} + \method + TT and CLIP~\cite{Radford2021} on the \hicodet \publicsetting setting. Our proposed \method surpasses the CLIP baseline in predicting interactions.}
    \label{fig:supp_qualitatives}
\end{figure*}

\begin{figure*}
    \centering
    \includegraphics[width=1.0\linewidth]{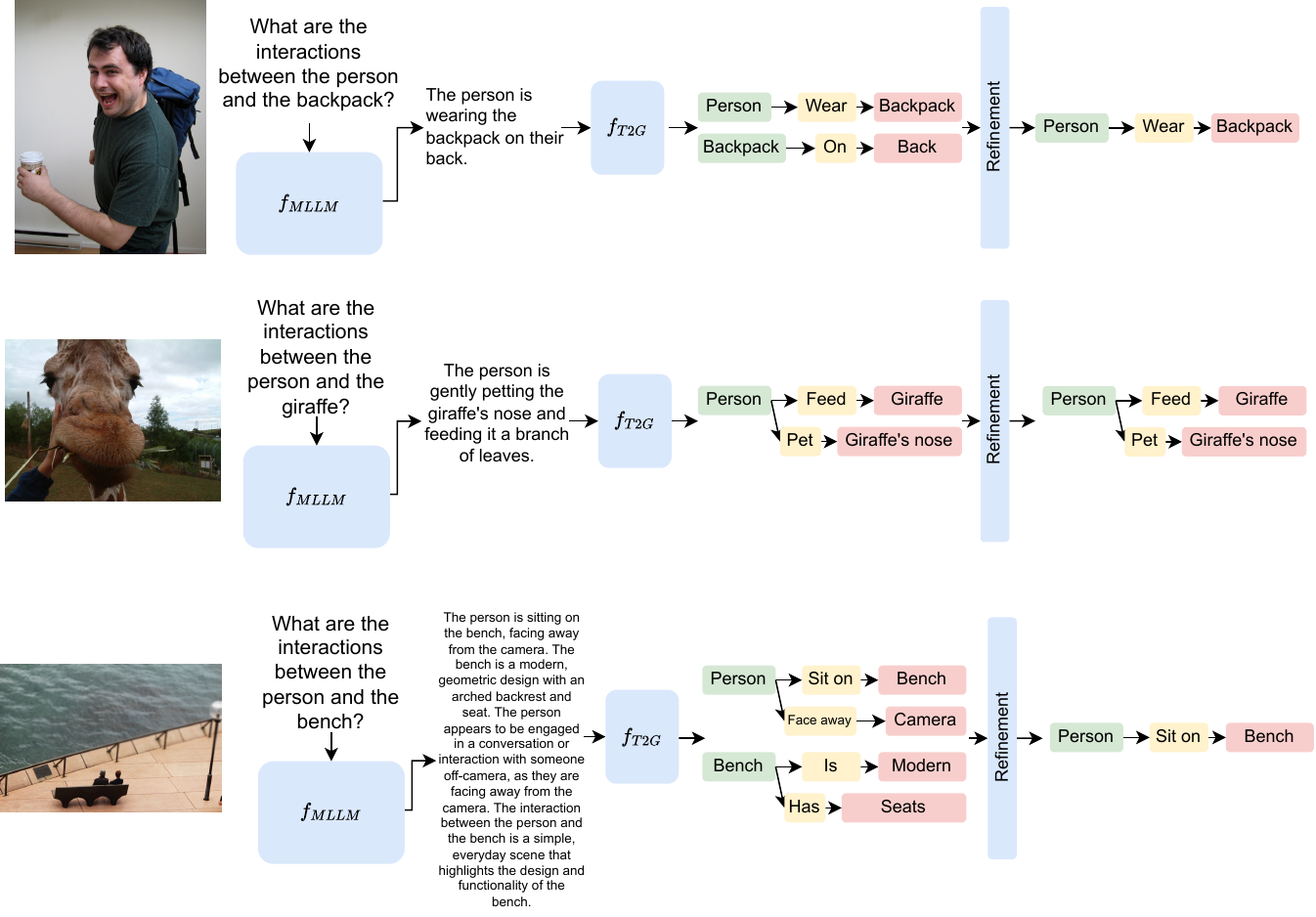}
    \caption{Qualitative results of LLaVA OV 0.5B~\cite{li2024llava} + \method + TT, and the intermediate outputs, on the \hicodet \publicsetting setting.}
    \label{fig:supp_qualitatives_big}
\end{figure*}

In \cref{fig:supp_qualitatives,fig:supp_qualitatives_big} we provide additional qualitative results of LLAVA OV 0.5B~\cite{li2024llava} + \method + TT and CLIP~\cite{Radford2021}.

\section{Ethics statement}
\label{sec:ethics}
We do not anticipate any immediate negative societal impact from our work. However, we urge future researchers to recognize that advances in automated human-object interaction detection may be applied for both beneficial and potentially harmful purposes.

The primary objective of our research is to enhance real-world human-object interaction detection by eliminating the constraints of predefined verb vocabularies. Our approach leverages established, publicly available datasets (e.g., \hicodet and \vghoi) and strictly adheres to their licensing agreements. Importantly, our work does not involve the collection of new images or sensitive personal data, and all experiments are conducted on local systems, thereby ensuring that user and institutional privacy are maintained.

While we believe that our contributions can significantly improve the understanding of dynamic visual scenes, we caution that such technologies might be repurposed in applications such as surveillance or behavior monitoring. We therefore encourage future work to carefully consider ethical implications and to employ these methods responsibly, with a commitment to transparency and adherence to ethical guidelines.

\newpage
\clearpage

\newpage
\clearpage

\begin{table*}[!t]
\caption{Performance on the \hicodet dataset for \task under the \publicsetting setting, using different thresholds for mAP calculation.}
\centering
\resizebox{1.0\linewidth}{!}{
\begin{tabular}{cc|ccc|ccc|ccc|ccc|ccc}
\toprule
\multicolumn{2}{c|}{\multirow{2}{*}{\textbf{Method}}}  & \multicolumn{3}{c|}{\textbf{mAP @0.6 (\%)} $\uparrow$} & \multicolumn{3}{c|}{\textbf{mAP @0.7 (\%)} $\uparrow$} & \multicolumn{3}{c|}{\textbf{mAP @0.8 (\%)} $\uparrow$} & \multicolumn{3}{c|}{\textbf{mAP @0.9 (\%)} $\uparrow$} & \multicolumn{3}{c}{\textbf{mAP @0.95 (\%)} $\uparrow$}  \\
&& \multicolumn{1}{c}{Full} & \multicolumn{1}{c}{Rare} & \multicolumn{1}{c|}{Non-rare} & \multicolumn{1}{c}{Full} & \multicolumn{1}{c}{Rare} & \multicolumn{1}{c|}{Non-rare} & \multicolumn{1}{c}{Full} & \multicolumn{1}{c}{Rare} & \multicolumn{1}{c|}{Non-rare} & \multicolumn{1}{c}{Full} & \multicolumn{1}{c}{Rare} & \multicolumn{1}{c|}{Non-rare} & \multicolumn{1}{c}{Full} & \multicolumn{1}{c}{Rare} & \multicolumn{1}{c}{Non-rare} \\
\midrule
\rowcolor{ourtablecolor} {}
 & VG-HOI & 9.69 & 16.24 & 7.73 & 8.14 & 17.05 & 5.48 & 5.86 & 15.93 & 2.85 & 2.78 & 8.14 & 1.18 & 0.00 & 0.00 & 0.00 \\
\rowcolor{ourtablecolor} {\multirow{-2}{*}{{CLIP~\cite{Radford2021}}}} & WordNet & 7.34 & 16.46 & 4.61 & 6.88 & 16.51 & 4.00 & 5.83 & 15.54 & 2.92 & 2.52 & 7.73 & 0.97 & 0.00 & 0.00 & 0.00  \\
\multicolumn{2}{l|}{CogVLM2 19B~\cite{hong2024cogvlm2}} & 3.03 & 2.10 & 3.30 & 2.67 & 2.10 & 2.84 & 1.73 & 1.37 & 1.84 & 1.51 & 1.15 & 1.61 & 1.49 & 1.15 & 1.59  \\
\multicolumn{2}{l|}{\textbf{+ \method (Ours)}} & 14.27 & 12.78 & 14.72 & 13.51 & 12.64 & 13.77 & 12.55 & 12.51 & 12.56 & 12.39 & 12.17 & 12.45 & 12.34 & 12.17 & 12.39  \\ 
\rowcolor{ourtablecolor} \multicolumn{2}{l|}{Idefics2 8B~\cite{laurenccon2025matters}} & 8.19 & 7.28 & 8.46 & 7.85 & 6.89 & 8.14 & 6.07 & 5.50 & 6.24 & 5.46 & 5.11 & 5.57 & 5.40 & 5.11 & 5.49 \\
\rowcolor{ourtablecolor} \multicolumn{2}{l|}{\textbf{+ \method (Ours)}} & 11.16 & 11.02 & 11.20 & 10.52 & 10.56 & 10.50 & 10.16 & 9.99 & 10.21 & 9.82 & 9.59 & 9.89 & 9.76 & 9.59 & 9.81 \\ 
\multicolumn{2}{l|}{InternVL2 2B~\cite{chen2024internvl}} & 3.30 & 1.84 & 3.74 & 3.25 & 1.84 & 3.67 & 2.84 & 1.58 & 3.22 & 2.62 & 1.58 & 2.92 & 2.58 & 1.58 & 2.87  \\
\multicolumn{2}{l|}{\textbf{+ \method (Ours)}} & 11.50 & 9.70 & 12.04 & 10.97 & 9.39 & 11.45 & 9.62 & 8.46 & 9.97 & 9.50 & 8.40 & 9.83 & 9.46 & 8.40 & 9.77  \\ 
\rowcolor{ourtablecolor} \multicolumn{2}{l|}{InternVL2 4B~\cite{chen2024internvl}} & 4.37 & 1.98 & 5.08 & 4.31 & 1.98 & 5.00 & 3.05 & 1.54 & 3.50 & 2.73 & 1.54 & 3.09 & 2.71 & 1.54 & 3.07  \\
\rowcolor{ourtablecolor} \multicolumn{2}{l|}{\textbf{+ \method (Ours)}} & 11.20 & 9.35 & 11.75 & 10.72 & 9.02 & 11.23 & 9.42 & 7.96 & 9.86 & 9.26 & 7.69 & 9.72 & 9.21 & 7.69 & 9.67  \\ 
\multicolumn{2}{l|}{InternVL2 8B~\cite{chen2024internvl}} & 3.55 & 1.48 & 4.16 & 3.44 & 1.35 & 4.06 & 2.60 & 1.28 & 2.99 & 2.43 & 1.28 & 2.78 & 2.41 & 1.28 & 2.75  \\
\multicolumn{2}{l|}{\textbf{+ \method (Ours)}} & 12.27 & 11.10 & 12.61 & 11.87 & 10.73 & 12.21 & 10.78 & 10.47 & 10.88 & 10.70 & 10.54 & 10.75 & 10.64 & 10.54 & 10.67  \\ 
\rowcolor{ourtablecolor} \multicolumn{2}{l|}{InstructBLIP 7B~\cite{huang2023visual}} & 4.75 & 4.47 & 4.84 & 4.70 & 4.44 & 4.78 & 4.01 & 3.62 & 4.13 & 3.82 & 3.62 & 3.88 & 3.74 & 3.62 & 3.78  \\
\rowcolor{ourtablecolor} \multicolumn{2}{l|}{\textbf{+ \method (Ours)}} & 13.34 & 14.72 & 12.93 & 12.78 & 14.02 & 12.41 & 10.65 & 12.29 & 10.16 & 10.46 & 12.02 & 9.99 & 10.43 & 12.02 & 9.95  \\ 
\multicolumn{2}{l|}{InstructBLIP 13B~\cite{huang2023visual}} & 2.52 & 0.58 & 3.10 & 2.50 & 0.58 & 3.08 & 1.70 & 0.30 & 2.12 & 1.45 & 0.20 & 1.82 & 1.43 & 0.20 & 1.79  \\
\multicolumn{2}{l|}{\textbf{+ \method (Ours)}} & 12.91 & 14.58 & 12.41 & 12.59 & 14.10 & 12.14 & 10.10 & 12.03 & 9.53 & 9.96 & 11.96 & 9.36 & 9.89 & 11.96 & 9.27   \\ 
\rowcolor{ourtablecolor} \multicolumn{2}{l|}{Phi3V 4B~\cite{abdin2024phi}} & 2.28 & 1.24 & 2.60 & 2.26 & 1.24 & 2.56 & 2.14 & 1.24 & 2.41 & 1.85 & 1.21 & 2.05 & 1.82 & 1.21 & 2.00  \\
\rowcolor{ourtablecolor} \multicolumn{2}{l|}{\textbf{+ \method (Ours)}} & 11.87 & 11.62 & 11.95 & 11.29 & 10.83 & 11.43 & 10.38 & 10.29 & 10.41 & 10.21 & 10.24 & 10.20 & 10.16 & 10.24 & 10.13  \\ 
\multicolumn{2}{l|}{Qwen2-VL 2B~\cite{wang2024qwen2}} & 3.80 & 2.26 & 4.26 & 3.77 & 2.26 & 4.22 & 3.26 & 2.20 & 3.57 & 2.99 & 2.17 & 3.23 & 2.95 & 2.17 & 3.19 \\
\multicolumn{2}{l|}{\textbf{+ \method (Ours)}} & 14.48 & 15.46 & 14.19 & 13.90 & 14.39 & 13.76 & 12.23 & 12.71 & 12.08 & 11.92 & 12.35 & 11.79 & 11.89 & 12.35 & 11.76  \\ 
\rowcolor{ourtablecolor} \multicolumn{2}{l|}{Qwen2-VL 7B~\cite{wang2024qwen2}} & 7.49 & 8.40 & 7.21 & 7.40 & 8.40 & 7.10 & 6.14 & 7.66 & 5.68 & 5.74 & 7.63 & 5.17 & 5.67 & 7.63 & 5.08 \\
\rowcolor{ourtablecolor} \multicolumn{2}{l|}{\textbf{+ \method (Ours)}} & 15.74 & 18.12 & 15.03 & 15.34 & 17.69 & 14.65 & 14.19 & 16.39 & 13.54 & 14.09 & 16.19 & 13.47 & 14.05 & 16.19 & 13.41 \\ 
\multicolumn{2}{l|}{LLaVA OV 7B~\cite{li2024llava}} & 4.78 & 5.34 & 4.61 & 4.71 & 5.34 & 4.52 & 4.22 & 4.68 & 4.08 & 3.80 & 4.46 & 3.60 & 3.77 & 4.46 & 3.57  \\
\multicolumn{2}{l|}{\textbf{+ \method (Ours)}} & 13.96 & 15.37 & 13.53 & 13.49 & 15.12 & 13.01 & 12.62 & 13.44 & 12.37 & 12.52 & 13.33 & 12.27 & 12.48 & 13.33 & 12.23  \\
\rowcolor{ourtablecolor} \multicolumn{2}{l|}{LLaVA OV 0.5B~\cite{li2024llava}} & 1.69 & 0.26 & 2.12 & 1.56 & 0.26 & 1.94 & 1.35 & 0.07 & 1.73 & 1.02 & 0.07 & 1.31 & 0.99 & 0.07 & 1.27 \\
\rowcolor{ourtablecolor}  \multicolumn{2}{l|}{\textbf{+ \method (Ours)}} & 14.83 & 16.02 & 14.47 & 14.18 & 15.56 & 13.77 & 12.92 & 15.06 & 12.28 & 12.71 & 14.48 & 12.18 & 12.66 & 14.48 & 12.12  \\
\rowcolor{ourtablecolor}  \multicolumn{2}{l|}{\textbf{+ \method + TT (Ours)}} & 23.19 & 25.79 & 22.41 & 22.64 & 25.51 & 21.78 & 19.41 & 24.90 & 17.76 & 19.28 & 25.02 & 17.56 & 19.23 & 25.02 & 17.50 \\ 
\bottomrule
\end{tabular}}
\label{tab:supp_hicodet_classification}
\end{table*}

\begin{table*}[!t]
\caption{Performance on the \hicodet dataset for \task under the \privatesetting setting, using different thresholds for mAP calculation}
\centering
\resizebox{1.0\linewidth}{!}{
\begin{tabular}{lc|ccc|ccc|ccc|ccc|ccc}
\toprule
\multicolumn{2}{c|}{\multirow{2}{*}{\textbf{Method}}}  & \multicolumn{3}{c|}{\textbf{mAP @0.6 (\%)} $\uparrow$} & \multicolumn{3}{c|}{\textbf{mAP @0.7 (\%)} $\uparrow$} & \multicolumn{3}{c|}{\textbf{mAP @0.8 (\%)} $\uparrow$} & \multicolumn{3}{c|}{\textbf{mAP @0.9 (\%)} $\uparrow$} & \multicolumn{3}{c}{\textbf{mAP @0.95 (\%)} $\uparrow$}  \\
&& \multicolumn{1}{c}{Full} & \multicolumn{1}{c}{Rare} & \multicolumn{1}{c|}{Non-rare} & \multicolumn{1}{c}{Full} & \multicolumn{1}{c}{Rare} & \multicolumn{1}{c|}{Non-rare} & \multicolumn{1}{c}{Full} & \multicolumn{1}{c}{Rare} & \multicolumn{1}{c|}{Non-rare} & \multicolumn{1}{c}{Full} & \multicolumn{1}{c}{Rare} & \multicolumn{1}{c|}{Non-rare} & \multicolumn{1}{c}{Full} & \multicolumn{1}{c}{Rare} & \multicolumn{1}{c}{Non-rare} \\
\midrule
\rowcolor{ourtablecolor} {}
 & VG-HOI & 6.36 & 10.71 & 5.06 & 5.47 & 11.39 & 3.70 & 3.54 & 9.37 & 1.79 & 1.85 & 5.61 & 0.73 & 0.00 & 0.00 & 0.00 \\
\rowcolor{ourtablecolor} {\multirow{-2}{*}{{CLIP~\cite{Radford2021}}}} & WordNet & 4.86 & 11.51 & 2.88 & 4.72 & 11.56 & 2.68 & 4.12 & 10.67 & 2.16 & 1.94 & 5.77 & 0.80 & 0.00 & 0.00 & 0.00  \\
 \multirow{2}{*}{{GEN-VLKT$_\text{CVPR22}$~\cite{Liao2022}}} & VG-HOI & 8.18 & 7.67 & 8.34 & 6.07 & 7.13 & 5.75 & 4.90 & 6.20 & 4.52 & 3.00 & 2.43 & 3.18 & 2.82 & 2.51 & 2.91 \\
& WordNet & 6.28 & 4.76 & 6.73 & 4.65 & 2.48 & 5.29 & 3.37 & 1.74 & 3.86 & 2.14 & 0.73 & 2.56 & 1.91 & 0.73 & 2.27 \\

\rowcolor{ourtablecolor} & VG-HOI & 1.67 & 2.66 & 1.37 & 1.56 & 2.38 & 1.31 & 1.46 & 1.9 & 1.32 & 1.48 & 1.91 & 1.35 & 1.48 & 1.91 & 1.35 \\
\rowcolor{ourtablecolor} \multirow{-2}{*}{{HOICLIP$_\text{CVPR23}$~\cite{Ning2023}}} & WordNet & 2.71 & 2.10 & 2.89 & 2.84 & 2.33 & 2.99 & 2.79 & 2.37 & 2.91 & 2.76 & 2.24 & 2.91 & 2.72 & 2.32 & 2.84 \\

& VG-HOI & 3.24 & 1.47 & 3.76 & 2.38 & 1.01 & 2.79 & 2.17 & 0.74 & 2.6 & 1.57 & 0.57 & 1.87 & 1.47 & 0.57 & 1.74 \\
\multirow{-2}{*}{{LogicHOI$_\text{NeurIPS23}$~\cite{li2023neural}}} & WordNet & 1.68 & 0.70 & 1.97 & 1.59 & 0.70 & 1.86 & 1.4 & 0.32 & 1.72 & 1.39 & 0.32 & 1.71 & 1.39 & 0.32 & 1.71 \\

\rowcolor{ourtablecolor} & VG-HOI & 8.48 & 12.06 & 7.42 & 7.88 & 11.93 & 6.67 & 6.84 & 11.65 & 5.41 & 5.46 & 10.56 & 3.94 & 5.11 & 10.25 & 3.58 \\
\rowcolor{ourtablecolor} \multirow{-2}{*}{{MP-HOI$_\text{CVPR24}$~\cite{Yang24}}} & WordNet & 7.64 & 9.74 & 7.01 & 6.54 & 8.95 & 5.82 & 5.05 & 7.46 & 4.33 & 4.11 & 7.58 & 3.08 & 4.03 & 7.56 & 2.97 \\ 

& VG-HOI & 8.77 & 3.96 & 10.21 & 8.22 & 3.52 & 9.62 & 7.38 & 3.24 & 8.62 & 5.61 & 1.84 & 6.74 & 4.89 & 1.05 & 6.03 \\
\multirow{-2}{*}{{LAIN$_\text{CVPR25}$~\cite{kim2025locality}}} & WordNet &  1.42 & 1.42 & 1.42 & 1.12 & 1.07 & 1.13 & 0.95 & 0.92 & 0.96 & 0.9 & 0.92 & 0.89 & 0.88 & 0.92 & 0.87 \\

\rowcolor{ourtablecolor} \multicolumn{2}{l|}{CogVLM2 19B~\cite{hong2024cogvlm2}} & 1.83 & 0.76 & 2.15 & 1.58 & 0.76 & 1.82 & 1.22 & 0.76 & 1.36 & 1.05 & 0.56 & 1.19 & 1.04 & 0.56 & 1.19 \\
\rowcolor{ourtablecolor} \multicolumn{2}{l|}{\textbf{+ \method (Ours)}} & 8.69 & 7.64 & 9.01 & 8.29 & 7.25 & 8.60 & 7.93 & 7.49 & 8.06 & 7.83 & 7.38 & 7.96 & 7.78 & 7.38 & 7.90  \\ 

\multicolumn{2}{l|}{InternVL2 8B~\cite{chen2024internvl}} & 2.25 & 1.28 & 2.54 & 2.17 & 1.28 & 2.44 & 1.81 & 1.15 & 2.01 & 1.68 & 1.15 & 1.84 & 1.65 & 1.15 & 1.80  \\
\multicolumn{2}{l|}{\textbf{+ \method (Ours)}} & 7.01 & 6.12 & 7.28 & 6.75 & 5.98 & 6.98 & 6.30 & 5.63 & 6.50 & 6.16 & 5.63 & 6.31 & 6.15 & 5.63 & 6.30  \\ 

\rowcolor{ourtablecolor} \multicolumn{2}{l|}{Qwen2-VL 7B~\cite{wang2024qwen2}} & 4.53 & 4.69 & 4.49 & 4.46 & 4.69 & 4.39 & 3.94 & 4.53 & 3.76 & 3.64 & 4.42 & 3.41 & 3.59 & 4.42 & 3.35 \\
\rowcolor{ourtablecolor} \multicolumn{2}{l|}{\textbf{+ \method (Ours)}} & 9.88 & 11.03 & 9.54 & 9.61 & 11.05 & 9.18 & 9.13 & 10.41 & 8.75 & 9.08 & 10.31 & 8.71 & 9.03 & 10.31 & 8.65 \\ 

\multicolumn{2}{l|}{LLaVA OV 7B~\cite{li2024llava}} & 3.00 & 2.58 & 3.12 & 2.95 & 2.58 & 3.05 & 2.73 & 2.46 & 2.82 & 2.48 & 2.42 & 2.50 & 2.46 & 2.42 & 2.47  \\
\multicolumn{2}{l|}{\textbf{+ \method (Ours)}} & 9.43 & 11.59 & 8.78 & 9.01 & 11.13 & 8.38 & 8.54 & 10.30 & 8.01 & 8.36 & 9.89 & 7.91 & 8.33 & 9.89 & 7.87  \\ 

\rowcolor{ourtablecolor} \multicolumn{2}{l|}{LLaVA OV 0.5B~\cite{li2024llava}} & 1.63 & 0.20 & 2.05 & 1.53 & 0.20 & 1.93 & 1.36 & 0.20 & 1.71 & 1.08 & 0.20 & 1.34 & 1.05 & 0.20 & 1.31 \\
\rowcolor{ourtablecolor} \multicolumn{2}{l|}{\textbf{+ \method (Ours)}} & 7.04 & 5.53 & 7.49 & 6.73 & 5.23 & 7.18 & 6.21 & 4.93 & 6.60 & 6.11 & 4.89 & 6.47 & 6.08 & 4.89 & 6.43  \\
\rowcolor{ourtablecolor} \multicolumn{2}{l|}{\textbf{+ \method + TT (Ours)}} & 10.81 & 11.76 & 10.52 & 10.54 & 11.31 & 10.31 & 10.01 & 11.06 & 9.70 & 10.10 & 11.30 & 9.75 & 10.03 & 11.30 & 9.66 \\ 
\bottomrule
\end{tabular}}
\label{tab:supp_hicodet_detection}
\end{table*}

\begin{table*}[!t]
\caption{Performance on the \vghoi dataset for \task under the \publicsetting setting, using different thresholds for mAP calculation.}
\centering
\resizebox{1.0\linewidth}{!}{%
\begin{tabular}{lc|c|c|c|c|c}
\hline
\multicolumn{2}{c|}{\textbf{Method}}                         & \textbf{mAP @0.6 (\%) $\uparrow$} & \textbf{mAP @0.7 (\%) $\uparrow$} & \textbf{mAP @0.8 (\%) $\uparrow$} & \textbf{mAP @0.9 (\%) $\uparrow$} & \textbf{mAP @0.95 (\%) $\uparrow$} \\ 
\hline
\rowcolor{ourtablecolor} \multicolumn{1}{c}{} & VG-HOI & 27.70 & 15.33 & 3.63 & 1.29 & 1.07 \\
\rowcolor{ourtablecolor} \multicolumn{1}{c}{\multirow{-2}{*}{CLIP~\cite{Radford2021}}}                            & WordNet    & 4.96 & 3.30 & 2.78 & 1.97 & 1.72 \\
\multicolumn{1}{c}{\multirow{2}{*}{DHD$_\text{AAAI24}$~\cite{Wu2024}}}           & VG-HOI     & 21.41 & 18.54 & 2.57 & 0.65 & 0.50 \\
\multicolumn{1}{c}{}           & WordNet     & 11.34 & 9.34 & 8.79 & 7.57 & 7.27 \\ 
\rowcolor{ourtablecolor} \multicolumn{2}{l|}{CogVLM2~\cite{hong2024cogvlm2}}          & 4.82 & 3.59 & 3.00 & 0.00 & 0.00 \\
\rowcolor{ourtablecolor} \multicolumn{2}{l|}{\textbf{+ \method (Ours)}}         & 32.66 & 28.99 & 26.02 & 2.45 & 0.02 \\ 
\multicolumn{2}{l|}{Idefics2~\cite{laurenccon2025matters}}   & 23.22 & 20.52 & 18.52 & 0.43 & 0.00 \\
\multicolumn{2}{l|}{\textbf{+ \method (Ours)}}   & 35.00 & 31.02 & 28.47 & 3.32 & 0.17 \\ 
\rowcolor{ourtablecolor} \multicolumn{2}{l|}{InternVL2 2B~\cite{chen2024internvl}}     & 12.09 & 10.36 & 8.83 & 0.21 & 0.00 \\
\rowcolor{ourtablecolor} \multicolumn{2}{l|}{\textbf{+ \method (Ours)}}     & 27.90 & 24.24 & 21.64 & 2.78 & 0.11 \\ 
\multicolumn{2}{l|}{InternVL2 4B~\cite{chen2024internvl}}     & 15.39 & 12.66 & 11.08 & 0.23 & 0.00 \\
\multicolumn{2}{l|}{\textbf{+ \method (Ours)}}     & 32.22 & 27.63 & 24.92 & 3.07 & 0.14 \\ 
\rowcolor{ourtablecolor} \multicolumn{2}{l|}{InternVL2 8B~\cite{chen2024internvl}}     & 19.34 & 16.90 & 15.38 & 0.23 & 0.00 \\
\rowcolor{ourtablecolor} \multicolumn{2}{l|}{\textbf{+ \method (Ours)}}     & 35.54 & 31.45 & 28.35 & 3.68 & 0.13 \\ 
\multicolumn{2}{l|}{InstructBLIP 7B~\cite{huang2023visual}}  & 4.70 & 3.87 & 3.02 & 0.29 & 0.00 \\
\multicolumn{2}{l|}{\textbf{+ \method (Ours)}}  & 28.22 & 25.10 & 22.60 & 2.47 & 0.15 \\ 
\rowcolor{ourtablecolor} \multicolumn{2}{l|}{InstructBLIP 13B~\cite{huang2023visual}} & 3.79 & 2.97 & 2.48 & 0.13 & 0.00 \\
\rowcolor{ourtablecolor} \multicolumn{2}{l|}{\textbf{+ \method (Ours)}} & 27.59 & 23.90 & 21.61 & 2.80 & 0.12 \\ 
\multicolumn{2}{l|}{Phi3V~\cite{abdin2024phi}}               & 3.55 & 2.98 & 2.37 & 0.04 & 0.00 \\ 
\multicolumn{2}{l|}{\textbf{+ \method (Ours)}}               & 31.57 & 27.25 & 24.69 & 3.20 & 0.15 \\ 
\rowcolor{ourtablecolor} \multicolumn{2}{l|}{Qwen2-VL 2B~\cite{wang2024qwen2}}        & 8.68 & 7.50 & 6.47 & 0.19 & 0.00 \\
\rowcolor{ourtablecolor} \multicolumn{2}{l|}{\textbf{+ \method (Ours)}} &       26.05 & 22.23 & 19.96 & 2.42 & 0.14 \\ 
\multicolumn{2}{l|}{Qwen2-VL 7B~\cite{wang2024qwen2}}        & 17.99 & 15.26 & 13.90 & 0.19 & 0.00 \\
\multicolumn{2}{l|}{\textbf{+ \method (Ours)}}        & 35.80 & 30.93 & 28.16 & 2.89 & 0.25 \\ 
\rowcolor{ourtablecolor} \multicolumn{2}{l|}{LLaVA OV 7B~\cite{li2024llava}}          & 19.79 & 16.99 & 15.66 & 0.28 & 0.00 \\
\rowcolor{ourtablecolor} \multicolumn{2}{l|}{\textbf{+ \method (Ours)}}          & 38.94 & 34.02 & 31.06 & 4.07 & 0.14 \\ 
\multicolumn{2}{l|}{LLaVA OV 0.5B~\cite{li2024llava}}        & 2.57 & 1.95 & 1.41 & 0.13 & 0.00 \\
\multicolumn{2}{l|}{\textbf{+ \method (Ours)}}        & 23.24 & 19.86 & 17.79 & 2.77 & 0.13 \\
\multicolumn{2}{l|}{\textbf{+ \method + TT (Ours)}} & 43.54 & 38.77 & 33.84 & 5.97 & 0.17 \\
\bottomrule
\end{tabular}%
}
\label{tab:supp_vghoi_classification}
\end{table*}

\begin{table*}[!t]
\caption{Performance on the \vghoi dataset for \task under the \privatesetting setting, using different thresholds for mAP calculation.}
\centering
\resizebox{1.0\linewidth}{!}{%
\begin{tabular}{lc|c|c|c|c|c}
\hline
\multicolumn{2}{c|}{\textbf{Method}}                         & \textbf{mAP @0.6 (\%) $\uparrow$} & \textbf{mAP @0.7 (\%) $\uparrow$} & \textbf{mAP @0.8 (\%) $\uparrow$} & \textbf{mAP @0.9 (\%) $\uparrow$} & \textbf{mAP @0.95 (\%) $\uparrow$} \\
\hline
\rowcolor{ourtablecolor} \multicolumn{1}{c}{} & VG-HOI & 3.54 & 2.30 & 0.77 & 0.38 & 0.27 \\
\rowcolor{ourtablecolor} \multicolumn{1}{c}{\multirow{-2}{*}{CLIP~\cite{Radford2021}}}                            & WordNet    & 0.65 & 0.39 & 0.37 & 0.30 & 0.14 \\
\multicolumn{1}{c}{\multirow{2}{*}{DHD~\cite{Wu2024}}}           & VG-HOI     & 2.16 & 1.91 & 0.74 & 0.26 & 0.23 \\
\multicolumn{1}{c}{}           & WordNet     & 2.19 & 2.17 & 2.05 & 1.76 & 1.76 \\ 
\rowcolor{ourtablecolor} \multicolumn{2}{l|}{CogVLM2~\cite{hong2024cogvlm2}}          & 0.53 & 0.40 & 0.19 & 0.00 & 0.00 \\
\rowcolor{ourtablecolor} \multicolumn{2}{l|}{\textbf{+ \method (Ours)}}         & 4.04 & 3.34 & 2.55 & 0.61 & 0.00 \\ 
\multicolumn{2}{l|}{Idefics2~\cite{laurenccon2025matters}}   & 2.27 & 1.80 & 1.53 & 0.10 & 0.00 \\
\multicolumn{2}{l|}{\textbf{+ \method (Ours)}}   & 3.71 & 2.99 & 2.42 & 0.34 & 0.06 \\ 
\rowcolor{ourtablecolor} \multicolumn{2}{l|}{InternVL2 2B~\cite{chen2024internvl}}     & 2.22 & 1.70 & 1.35 & 0.10 & 0.00 \\
\rowcolor{ourtablecolor} \multicolumn{2}{l|}{\textbf{+ \method (Ours)}}     & 4.60 & 3.86 & 2.87 & 0.82 & 0.00 \\ 
\multicolumn{2}{l|}{InternVL2 4B~\cite{chen2024internvl}}     & 1.72 & 1.19 & 0.93 & 0.01 & 0.00 \\
\multicolumn{2}{l|}{\textbf{+ \method (Ours)}}     & 4.18 & 3.45 & 2.88 & 0.73 & 0.02 \\ 
\rowcolor{ourtablecolor} \multicolumn{2}{l|}{InternVL2 8B~\cite{chen2024internvl}}     & 2.08 & 1.54 & 1.39 & 0.16 & 0.00 \\
\rowcolor{ourtablecolor} \multicolumn{2}{l|}{\textbf{+ \method (Ours)}}     & 5.04 & 4.19 & 3.56 & 1.00 & 0.09 \\ 
\multicolumn{2}{l|}{InstructBLIP 7B~\cite{huang2023visual}}  & 0.71 & 0.61 & 0.51 & 0.08 & 0.00 \\
\multicolumn{2}{l|}{\textbf{+ \method (Ours)}}  & 4.86 & 4.11 & 3.20 & 0.85 & 0.06 \\ 
\rowcolor{ourtablecolor} \multicolumn{2}{l|}{InstructBLIP 13B~\cite{huang2023visual}} & 0.72 & 0.52 & 0.43 & 0.01 & 0.00 \\
\rowcolor{ourtablecolor} \multicolumn{2}{l|}{\textbf{+ \method (Ours)}} & 4.64 & 3.61 & 3.13 & 0.75 & 0.06 \\ 
\multicolumn{2}{l|}{Phi3V~\cite{abdin2024phi}}               & 0.53 & 0.31 & 0.28 & 0.01 & 0.00 \\ 
\multicolumn{2}{l|}{\textbf{+ \method (Ours)}}               & 3.64 & 2.83 & 2.32 & 0.45 & 0.09 \\ 
\rowcolor{ourtablecolor} \multicolumn{2}{l|}{Qwen2-VL 2B~\cite{wang2024qwen2}}        & 1.13 & 0.86 & 0.64 & 0.13 & 0.00 \\
\rowcolor{ourtablecolor} \multicolumn{2}{l|}{\textbf{+ \method (Ours)}} &       4.11 & 3.24 & 2.64 & 0.68 & 0.06 \\ 
\multicolumn{2}{l|}{Qwen2-VL 7B~\cite{wang2024qwen2}}        & 1.71 & 1.43 & 1.12 & 0.11 & 0.00 \\
\multicolumn{2}{l|}{\textbf{+ \method (Ours)}}        & 5.09 & 4.21 & 3.54 & 0.94 & 0.02 \\ 
\rowcolor{ourtablecolor} \multicolumn{2}{l|}{LLaVA OV 7B~\cite{li2024llava}}          & 1.91 & 1.47 & 1.27 & 0.03 & 0.00 \\
\rowcolor{ourtablecolor} \multicolumn{2}{l|}{\textbf{+ \method (Ours)}}          & 4.88 & 3.97 & 3.40 & 0.96 & 0.00 \\ 
\multicolumn{2}{l|}{LLaVA OV 0.5B~\cite{li2024llava}}        & 0.31 & 0.26 & 0.15 & 0.04 & 0.00 \\
\multicolumn{2}{l|}{\textbf{+ \method (Ours)}}        & 3.69 & 3.12 & 2.50 & 0.64 & 0.00 \\
\multicolumn{2}{l|}{\textbf{+ \method + TT (Ours)}} & 6.50 & 5.27 & 4.09 & 1.47 & 0.02 \\
\bottomrule
\end{tabular}%
}
\label{tab:supp_vghoi_detection}
\end{table*}

\newpage
\clearpage

\end{document}